\documentclass[10pt]{asme2ej}

\usepackage{epsfig} 

\usepackage{array}
\usepackage{amssymb}
\usepackage{amsmath}
\usepackage[thinc]{esdiff}
\usepackage{xcolor}
\usepackage{graphicx}
\usepackage{subcaption}
\usepackage{makecell}
\usepackage{float}

\title{Multicriteria Optimization of \\
Lower Limb Exoskeleton Mechanism}

\author{Sayat Ibrayev
    \affiliation{
	Professor\\
	Joldasbekov Institute of\\ 
    Mechanics and Engineering\\
	 Almaty, Kazakhstan, 050010\\
    Email: sayat\_m.ibrayev@mail.ru
    }	
}

\author{Arman Ibrayeva\thanks{Corresponding author}
    \affiliation{Senior Research Scientist\\
	Joldasbekov Institute of\\ 
    Mechanics and Engineering\\
	 Almaty, Kazakhstan, 050010\\
        Email: arman.ibrayeva@kaust.edu.sa
    }
}

\author{Ayaulym Rakhmatullina
        \affiliation{Associate Professor\\
        Department of Engineering Graphics\\
        and Applied Mechanics\\
        Almaty Technological University\\
        Almaty, Kazakhstan, 050012}
        }
\author{Aizhan Ibrayeva\\
    \affiliation{Scientist\\
    Megalabs, Palo Alto, CA 94301}
    }	

\author{Bekzat Amanov
    \affiliation{Senior Research Scientist\\
	Joldasbekov Institute of\\
 Mechanics and Engineering\\
	 Almaty, Kazakhstan, 050010\\
    }	
}
\author{Nurbibi Imanbayeva
        \affiliation{Associate Professor\\
        Department of Mechanical Engineering\\
        Satbayev University\\
        Almaty, Kazakhstan, 050000\\}
        }
        
\begin{document}

\maketitle    

\begin{abstract}
Typical leg exoskeletons employ open-loop kinematic chains with motors placed directly on movable joints;  while this design offers flexibility, it leads to increased costs and heightened control complexity due to the high number of degrees of freedom. The use of heavy servo-motors to handle torque in active joints results in complex and bulky designs, as highlighted in existing literature. In this study, we introduced a novel synthesis method with analytical solutions provided for synthesizing lower-limb exoskeleton. Additionally, we have incorporated multicriteria optimization by six designing criteria. As a result, we offer several mechanisms, comprising only six links, well-suited to the human anatomical structure, exhibit superior trajectory accuracy, efficient force transmission, satisfactory step height, and having internal transfer segment of the foot.
\end{abstract}


\section{Introduction}
\label{sec:intro}
Exoskeleton robots have found broad application for augmenting power and aiding in rehabilitation \cite{mikolajczyk2022recent, shi2019review,tijjani2022survey, sanchez2019compliant, copilusi2022design, shen2019customized, tarnita2018numerical}.  Power augmentation is important for tasks involving heavy load transportation with limited muscle strength, while robot-assisted technologies employing upper and lower limb exoskeletons are used for rehabilitating individuals who have experienced a loss of mobility in their joints and muscles. Studying human walking apparatus and motion diagrams representing the leg movement were useful in various fields including the development of human prosthetics, human mimicking robots, and advancements in research areas such as biomimetics, military combat, cinematography, toys, and terrestrial and extraterrestrial exploration\cite{tarnita2018numerical, humangait}. For a comprehensive overview of bipedal walking robots and exoskeletons, refer to \cite{mikolajczyk2022recent, shi2019review,tijjani2022survey, sanchez2019compliant}.

Various design and control architectures of exoskeletons were summarized in the relevant references \cite{copilusi2022design, tijjani2022survey}. In recent years, various design schemes of lower limb exoskeletons aimed at achieving compact devices and meeting specific optimality criteria have been proposed \cite{ishmael2022powered, geonea2019design, komoda2017energy}. Commonly, the kinematic scheme of the leg exoskeleton is based on an open-loop kinematic chain with the motors mounted directly on the movable joints. While these open kinematic chains offer greater flexibility and ease of design, their large number of degrees of freedom (DOF) contributes to increased costs and complexities in control. Using heavy servo-motors to meet significant torques generated in active joints leads to complicated and cumbersome design. The bulkiness and substantial weight of this kind of devices are highlighted in \cite{shen2019customized, shen2018integrated}.

 Numerous researchers have studied the walking apparatus, demonstrating that walking patterns are measurable, predictable, and repeatable. \cite{liu2022mechanical} presents a passive exoskeleton with 17 DOF for load-carrying, which includes two 3 DOF ankle joints, two 2 DOF hip joints, two 1 DOF knee joints, a 1 DOF backpack, and two redundant degrees of freedom at the thigh and the shank to improve the compatibility of human-machine locomotion. In \cite{terefe2019review}, a mechanism has been designed for a walking robot, effectively minimizing the number of required motors, thereby contributing to reduced energy consumption. The dimensional synthesis is conducted analytically to formulate a parametric equation, and the resulting geometry of the leg mechanism is established. However, it should be noted that this mechanism is not suitable for exoskeletons due to its significant deviation in shape from that of the human leg.
 
Shen et al. \cite{shen2018integrated, shen2019customized} introduced a 1 DOF mechanism for lower limb rehabilitation exoskeletons. In \cite{shen2018integrated} proposed an integrated type and dimensional synthesis method for designing compact 1 DOF planar linkages with only revolute joints, and applied the method on a leg exoskeleton that can generate human gait patterns simultaneously at hip and knee joints. Reducing the number of motors resulted in decreased energy consumption. \cite{shen2019customized} reports a close match to human gait and stable hip and knee joint outputs in their prototype due to its parallel structure. The drawback of Shen's mechanism is its use of an 8-bar-10-joint configuration, simplification would involve reducing the number of these linkages. 

\cite{geoneatarnita} presents a leg exoskeleton consisting of a planar five links closed linkage. The robotic system is intended for patients who have suffered strokes. The design is simple, wearable and light, with anthropomorphic structure, and operates with only one motor. The drawbacks mentioned in the paper include the limitation of providing mobility solely to the knee and hip joints, as well as the need for future design improvements, such as an adjustable length mechanism to accommodate patients with varying anthropometric data. In \cite{geonea2019design} the author introduced a new 1 DOF structure allowing adjustments of mechanism links to accommodate varying human body types. However, the geometry of the novel design is bulkier. 
The exoskeleton is based on seven-link mechanism, is cost-effective and easy to implement in practical activities. This leg mechanism can assure the mobility of knee and hip joints, and the ankle joint was not considered in favor of reducing the cost. Also the paper states that its design may require additional improvements in future developments, such as the foot shape optimization.
Another study \cite{tarnita2018numerical} focuses on the experimental and numerical study of human gait. Its results have practical applications in the design and development of human-inspired robotic structures for use in medical, assistive, or rehabilitation fields. Specifically, the study aims to investigate the flexion-extension movements of lower limb joints in humans and analyze the ground reaction forces generated during walking on force platforms.

Plecnik proposed a synthesis method for six-bar mechanisms \cite{plecnik2013dimensional}, which was applied to explore one million tasks, resulting in the synthesis of one hundred and twenty two practical linkage designs. In \cite{plecnik2016design} the author presents a design procedure for Stephenson I-III six-bar linkages that is demonstrated on the design of legs. These mechanisms are advantageous for their simplicity, characterized by a reduced number of links. While suitable for applications such as walking robots, they may not meet the requirements for exoskeletons, which demand designs that closely conform to the human anatomy while emphasizing compactness. Hence, the quest for more compact solutions in the context of exoskeletons continues.



Demonstration of a 1 DOF closed-loop mechanical linkage that can be designed to the shape and movement of a biped human walking apparatus is presented also in \cite{al2011conceptual}. A single DOF eight-bar path generator that typifies the shape and motion of a human leg is proposed. However, the relative foot stride is notably small. This concept was developed by the authors in \cite{batayneh2013biomimetic} towards minimizing the number of links, and a six-link leg mechanism for a biped robot was synthesized. Unfortunately, they used prismatic joints, and the foot stride is small as well. An eight-bar walking mechanism was designed in \cite{brown2006design}, but the foot path does not have a straight-line segment that will correspond to the support phase (when the foot contacts the ground).

Design and optimization of a 1 DOF eight-bar leg mechanism for a walking machine was proposed in \cite{Giesbrecht2010}. The leg mechanism is considered to be very energy efficient, especially when walking on rough terrains. Furthermore, the mechanism requires very simple controls since a single actuator is required to drive the leg. Dynamic analysis was performed to evaluate the joint forces and crank torques of each solution, thus taking into account inertia forces in the design. Two critical aspects of the leg mechanism's performance were chosen to be optimized: minimizing the energy to improve the efficiency of the leg and lowering the requirement for larger motors, and second, maximizing the stride length because a leg that travels a longer distance with lower energy is very desirable. Hence, the energy per unit of travel was reduced. However, the total design obtained was cumbersome, since the mechanism was based on Theo Jansen’s straight-line generator. 

The synthesis of leg mechanisms inspired from Theo-Jansen’s solution has been a research topic in the last years \cite{Punde2020, pop2016dimensional} for its advantages regarding the reduced number of DOFs which makes it easier to control, the scalable design, the reduced impact on the ground during walking, and because of less energy consumption. The Theo-Jansen mechanism was developed in \cite{Punde2020} considering an adaptive and controllable mechanism on irregular ground. This research sets a basis for further extension of the Theo-Jansen mechanism, considering the bending of the leg linkages while turning and providing high stiffness of design. The prototype developed was tested at various speeds and torques due to presence of the speed control system provided in electronic system, which enables to use the prototype for various load handling capacities. An eight-bar leg mechanism dimensional synthesis is presented in \cite{pop2016dimensional}  as well. As compared to the prototype, the advantage of the proposed geometric model consists of a greater step height that helps the robotic structure to overcome larger obstacles. 

The Klann linkage is another single DOF walking mechanism that is patented and is widely used on multi-legged robots \cite{kulandaidaasan2016trajectory,lokhande2013mechanical, soyguder2007design, komoda2012proposal}. The capabilities of standard non-reconfigurable quadruped Klann legs can be significantly extended applying the method proposed in \cite{kulandaidaasan2016trajectory}. Reconfigurable legged robots based on 1 DOF Klann mechanism are highly desired because they are effective on rough and irregular terrains and they provide mobility in such terrain with simple control schemes. However, both Theo Jansen’s solution and Klann mechanism inspired leg designs are very cumbersome, especially they cannot be used as lower limb exoskeleton since it does not fit the size limit and overall dimension restrictions. Other related works include \cite{kim2014optimal, xu2019kinematic, ibrayev2022optimization, ibrayev2019optimal, ibrayev2023optimal, ibrayev2020walking}.

Tsuge \cite{tsuge2015kinematics} proposed a kinematic synthesis method developed to achieve a mechanical system that guides a natural ankle trajectory for human walking gait. The author analyzed existing Watt I and Stephenson III six-bar linkage synthesis methods and applied the developed additional linkage synthesis procedures for treadmill training mechanism design. A new six-bar linkage system was proposed to support natural movement of the human lower limb. However the accuracy of generated straight-line segment of foot path is poor and relative horizontal velocity of the foot is not constant.

In \cite{desai2019analysis} presented 6 configurations of an 8-bar leg mechanism, with three fixed pivots that make it strong and stable, validated on experimental prototype. The paper emphasizes that this mechanism offers the largest stride-to-size ratio, allowing for the construction of a compact and lightweight walking mechanism with low inertia. Consequently, it is well-suited for speed walking. In this study we synthesized a mechanism that is even more compact, and comprising only six links. We introduced a novel synthesis method, and as the result of multicriteria optimization, we achieved a compact solution matching human anatomy, with high accuracy of trajectory generation,  optimized force transfer, minimized chassis height, and with internal transfer of foot. 

\begin{figure}[t]
    \centering
    \includegraphics[width=0.6\linewidth]{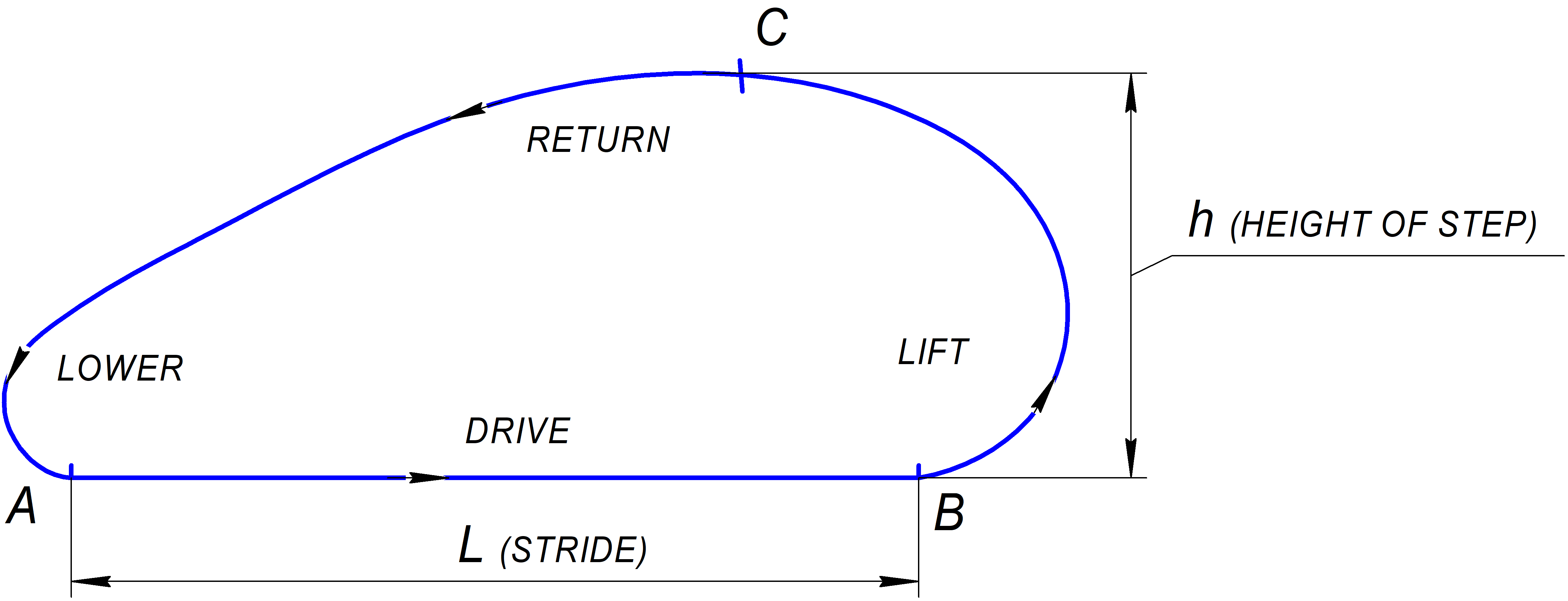}
    \caption{Prof. J.Shigley's Foot-Path Diagram}
    \label{fig:Sh_foot_path}
\end{figure}

The desired foot trajectory consists of two segments (Fig. \ref{fig:Sh_foot_path}): 
\begin{enumerate}
    \item straight-line segment $A - B$ with stride $ L$, corresponding to the \textit{support phase} of step cycle (when the foot $P$ is on the ground);
    \item swing phase segment $B - C - A$ with a step height $h$ (the leg \textit{transfer phase}).
\end{enumerate}

This paper is organized as follows: Section \ref{sec:str&kin} introduces the structure of the lower-limb exoskeleton mechanism and our proposed synthesis method. Section \ref{sec:analit_sln} presents analytical solutions of the synthesis problem. Multicriteria optimization conditions are defined in Section \ref{sec:additionalCrit}. Obtained design solutions are discussed in Section \ref{sec:globalsearch}, and the final design is presented in Section \ref{sec:finaldesign}. And Section \ref{sec:conclusions} provides conclusions.

 \section{Lower Limb Exoskeleton Mechanism Structure and Synthesis Problem Formulation}
 \label{sec:str&kin}
 

Human walking can be analyzed in three planes: the sagittal plane, coronal plane, and transverse plane \cite{shen2019customized, shen2018integrated}. Among these, the sagittal plane motion is predominant. The designed leg exoskeleton aids in hip and knee flexion/extension movements while the wearer stays in place. This practice focuses on motions within the sagittal plane. Consequently, a planar linkage with angular outputs can effectively facilitate these movements.

\begin{figure}[t]
    \centering
    \begin{subfigure}{0.3\textwidth}
        \centering
        \includegraphics[width=0.7\linewidth]{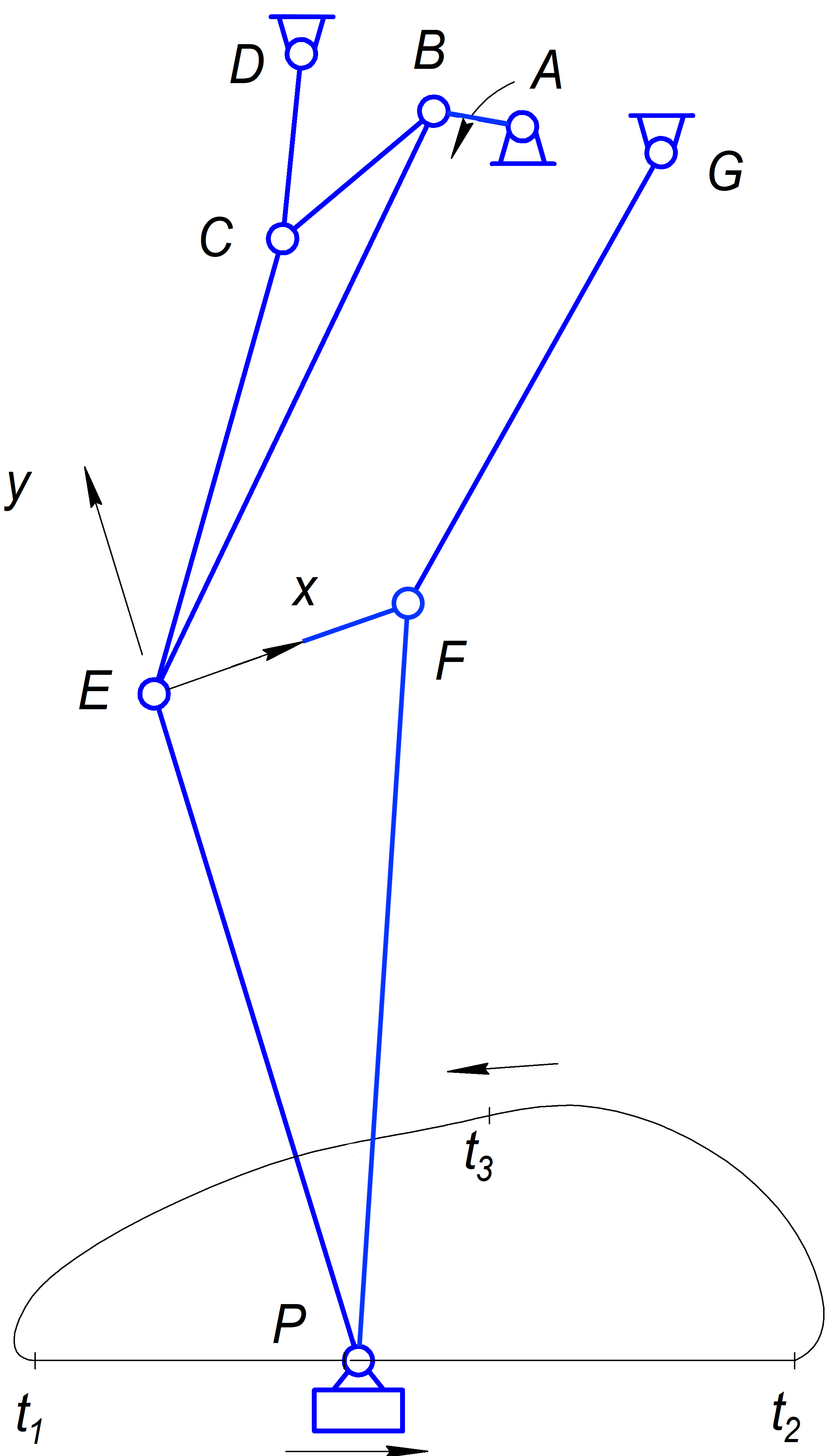}
        \caption{}
        \label{fig:1a}
    \end{subfigure}
    \begin{subfigure}{0.35\textwidth}
        \centering
        \includegraphics[width=0.7\linewidth]{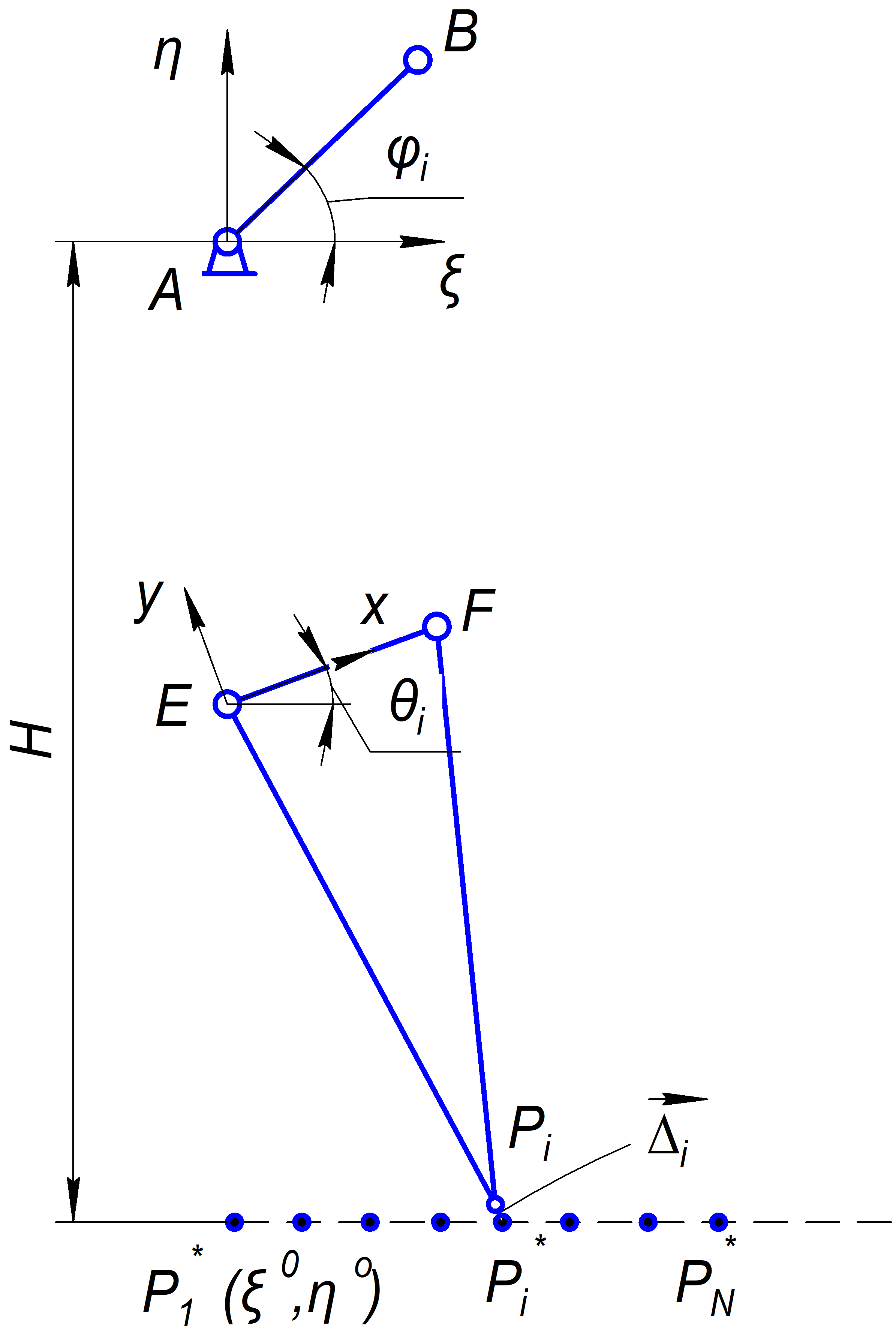}
        \caption{}
        \label{fig:1b}
    \end{subfigure}
    \hfill
    \caption{Kinematic scheme of a lower-limb exoskeleton mechanism}
    \label{fig:1}
\end{figure}

The kinematic scheme of six-bar Stephenson III type lower-limb exoskeleton mechanism is illustrated in Fig. \ref{fig:1a} with the input link $AB$ and the foot $P$ mounted on coupler $EF$. Rotation of the crank $AB$ is described by $\varphi_i$,

\begin{equation}
    \label{eq:1}
    \varphi_i = \varphi_0 + \Delta \Phi \frac{i-1}{N-1}, \quad i=\overline{1,N}.
\end{equation}

Here, $\varphi_0$ is an initial angular position of the crank $AB$ with respect to the horizontal axis. $\Delta \Phi$ is the maximum rotation angle, $\Delta \Phi > \pi$, in order to supply the overlap of support phases of alternating two legs. \textbf{Kinematic analysis} of the mechanism is provided in Appendix A. 

The rotation of the crank $GF$ is represented by the angle $\theta_i$, $i=\overline{1,N}$.
 $\vec{r}_P[x_P, y_P]$ is a local radius vector, rigidly associated with the moving coordinate system $Exy$; and $\vec{R}_{P_i}\left[\xi_{P_i},\eta_{P_i} \right]$ and $\vec{R}_{E_i}\left[\xi_{E_i},\eta_{E_i} \right]$ signify the absolute positions of the foot center $P_i$ and the point $E$ respectively. Then the relationship for $\vec{R}_{P_i}$ can be expressed as:

\begin{equation}
    \label{eq:2}
    \vec{R}_{P_i} = \vec{R}_{E_i} + \Gamma ( \theta_i ) \vec{r}_P
    = \begin{bmatrix}
        \xi_{E_i} \\
        \eta_{E_i}
      \end{bmatrix}
      + \begin{bmatrix}
        x_P \cos{\theta_i}-y_P \sin{\theta_i} \\
        x_P \sin{\theta_i}+y_P \cos{\theta_i}
      \end{bmatrix},
\end{equation}

\label{sec:eqs}
Given the desired absolute coordinates of the foot $P_i^*$ (Fig. \ref{fig:1b}): 
\[
\xi_{P_i}^*=\xi^0+L\frac{i-1}{N-1}, i=\overline{1,N}
\]

\[
\eta^*_{P_i}=\eta^0, 
\]
with stride $L$ and desired start position $P_1^*(\xi^0, \eta^0)$, the synthesis condition is stated as follows:

\begin{equation}
    \label{eq:4}
    \vec{\Delta}_i \equiv \vec{R}_{P_i}-\vec{R}_{P_i}^*=\vec{0}, \quad  i=\overline{1,N}.
\end{equation}

Considering the constraint equations Eq. (\ref{eq:2}) one can write in scalar form as follows:

\begin{equation}
\begin{aligned}
    \label{eq:7}
   \delta_i^{\xi}&=x_P \cos{\theta_i}-y_P \sin{\theta_i}-\xi^0+\xi_{E_i}-L\frac{i-1}{N-1}=0 \\
   \delta_i^{\eta}&=x_P \sin{\theta_i}+y_P \cos{\theta_i}-\eta^0+\eta_{E_i}=0.
\end{aligned}
\end{equation}
Given the mechanism link lengths except ($x_P, y_P$), synthesis task consists in determining the parameters $x_P, y_P, \xi^0, \eta^0, L$ that satisfy these constraints approximately. 

Then the least square approximation problem is formulated as follows:

\begin{equation}
    \label{eq:8}
    S=\sum_{i=1}^{N}\left( (\delta_i^\xi)^2+(\delta_i^\eta)^2 \right) \rightarrow \min_{\vec{x}}.
\end{equation}

\section{Analytical Solution of the Synthesis Problem }
\label{sec:analit_sln}
\subsection{4-parameter synthesis}
Unknown variables are: 

$x_1:=x_P, x_2:=y_P, x_3:=\xi^0, x_4:=\eta^0$.

The conditions $\frac{\partial S}{\partial {x_j}} = 0, j=\overline{1,4}$ lead to the following equations involving four unknowns $x_1, x_2, x_3, x_4$:



\begin{equation}
    \label{eq:9}
    \sum_{i=1}^{N} \left( \delta_i^{\xi} \cdot \frac{\partial \delta_i^\xi}{\partial x_j} + \delta_i^{\eta} \cdot \frac{\partial \delta_i^\eta}{\partial x_j} \right) = 0, \quad j = \overline{1,4}.
\end{equation}
Namely,

\begin{equation}
    \label{eq:10}
    \sum_{i=1}^{N} \delta_i^\xi \cos{\theta_i}+\sum_{i=1}^{N} \delta_i^\eta \sin{\theta_i}=0
\end{equation}

\begin{equation}
    \label{eq:11}
    -\sum_{i=1}^{N} \delta_i^\xi \sin{\theta_i}+\sum_{i=1}^{N} \delta_i^\eta \cos{\theta_i}=0
\end{equation}

\begin{equation}
    \label{eq:12}
    \sum_{i=1}^{N} \delta_i^\xi=0
\end{equation}

\begin{equation}
    \label{eq:13}
    \sum_{i=1}^{N} \delta_i^\eta=0
\end{equation}
    Let us introduce notations:

\begin{subequations}
\begin{gather}
    k := \sum_{i=1}^{N} \cos{\theta_i}, \label{eq:k} \\
    m := \sum_{i=1}^{N}\sin{\theta_i}, \label{eq:m} \\
    b_1 := \sum_{i=1}^{N} \left(-\xi_{E_i}+L\cdot\frac{i-1}{N-1}\right) \cos{\theta_i}-\sum_{i=1}^{N} \eta_{E_i} \sin{\theta_i}, \label{eq:4c} \\
    b_2 := \sum_{i=1}^{N} \left(\xi_{E_i}-L\cdot\frac{i-1}{N-1}\right) \sin{\theta_i}-\sum_{i=1}^{N} \eta_{E_i} \cos{\theta_i}, \label{eq:4d} \\
    b_3 := \sum_{i=1}^{N} \left(-\xi_{E_i}+L\cdot\frac{i-1}{N-1}\right), \label{eq:4e} \\
    b_4 := \sum_{i=1}^{N} (-\eta_{E_i}). \label{eq:4f}
\end{gather}
\end{subequations}

   Then  we come to the system of linear equations expressed as follows:
    
    \begin{subequations}
    \label{eq:14}
    \begin{align}
        & N x_1-k x_3-mx_4 =b_1 \label{eq:14.1}\\
        & N x_2 +m x_3-k x_4 =b_2\label{eq:14.2}\\
        & k x_1 - m x_2 -N x_3 =b_3\label{eq:14.3}\\
        & m x_1 +k x_2 -N x_4 =b_4 \label{eq:14.4}
    \end{align}
    \end{subequations}

    The only solution of this system in case $k^2+m^2 \neq N^2$ is

    \begin{equation}
        \label{eq:20}
        x_1 = \frac{-N b_1 +k b_3 + m b_4}{k^2+m^2-N^2}
    \end{equation}
    \begin{equation}
        \label{eq:21}
        x_2 = \frac{-N b_2 -m b_3 + k b_4}{k^2+m^2-N^2}
    \end{equation}
    \begin{equation}
        \label{eq:18}
        x_3=\frac{N b_3-k b_1+m b_2}{k^2+m^2-N^2}
    \end{equation}

    \begin{equation}
        \label{eq:19}
        x_4=\frac{N b_4-m b_1 -k b_2}{k^2+m^2-N^2}.
    \end{equation}

The necessary conditions also serve as sufficient conditions for attaining the minimum of the function $S$: the matrix 


\[
\frac{d^2S}{d\vec{x}^2} = 
\begin{bmatrix}
    N & 0 & -k & -m \\
    0 & N & m  & -k \\
    -k & m & N & 0 \\
    -m & -k &  0 & N\\
\end{bmatrix}
\]
is positive-definite (in case $k^2+m^2 \neq N^2$), since the determinants of all corner minors are non-negative:
\begin{equation}
\label{eq:*}
\begin{aligned}
& \Delta_1 = N > 0, \quad \Delta_2 = N^2 > 0, \quad \Delta_3 = N^3 + N (k^2 + m^2) > 0, \\
& \Delta_4 = (k^2 + m^2 - N^2)^2> 0. \\
\end{aligned}
\end{equation} 

\subsection{5-parameter synthesis}
Let's explore a scenario where $L = x_5$ is an additional variable subject to optimization. Referring to Eq. (\ref{eq:14}), from the condition $\frac{\partial S}{\partial \vec{x}} = \vec{0}$  we get
\begin{equation}
    \label{eq:22}
    N x_1-k x_3 - m x_4 - \sum_{i=1}^{N} \left(\frac{i-1}{N-1} \cos{\theta_i}\right) x_5 = b_5, 
\end{equation} 
where
\[
b_5 = \sum_{i=1}^{N}\left(-\xi_{P_i} \cdot \cos{\theta_i}-\eta_{P_i} \cdot\sin{\theta_i}\right)
\]
\begin{equation}
    \label{23}
    N x_2 +m x_3 - k x_4 +\sum_{i=1}^{N} \left(\frac{i-1}{N-1} \sin{\theta_i}\right) x_5 = b_6,
\end{equation}
where 
\[
b_6 = \sum_{i=1}^{N}\left(\xi_{P_i} \cdot \sin{\theta_i}-\eta_{P_i} \cdot\cos{\theta_i}\right)
\]

\begin{equation}
    \label{eq:24}
    k x_1-m x_2 - N x_3 - \sum_{i=1}^{N}\left(\frac{i-1}{N-1} \right)x_5=b_7,
\end{equation}
where
\[
b_7 = - \sum_{i=1}^{N} \xi_{P_i}
\]
\begin{equation}
    \label{eq:25}
    m x_1 + k x_2 - N x_4 = b_8,
\end{equation}
where
\[
b_8 = - \sum_{i=1}^{N} \eta_{P_i}.
=b_4.\]

The last equation is derived from Eq. (\ref{eq:9}), $j=5$:

\begin{equation}
\label{eq:26}
\begin{aligned}
& \sum\left(\frac{i-1}{N-1}\cos \theta_i\right) x_1 -\sum\left(\frac{i-1}{N-1}\sin \theta_i\right) x_2 \\
& -\sum\left(\frac{i-1}{N-1}\right) x_3 + \sum\left(\frac{i-1}{N-1} \xi_{P_i}\right) x_4 + \sum\left(\frac{i-1}{N-1}\right)^2 x_5 = 0.
\end{aligned}
\end{equation}

Finally, taking into account 
\begin{equation}
\label{eq:29}
    \sum\left(\frac{i-1}{N-1}\right) = N/2 
\end{equation}

and

\begin{equation}
\label{eq:29'}
    \sum\left(\frac{i-1}{N-1}\right)^2 = \frac{N (2 N -1)}{6 (N-1)} , 
\end{equation}

we obtain the following system of linear equations

\begin{equation}
    \label{eq:**}
    A \vec{x} = \vec{b},
\end{equation}

where 

\begin{equation}
    \label{eq:mtx_A}
    A = \begin{bmatrix}
       N        &  0      & -k & -m & -k_\alpha\\
       0        &  N      & m  & -k & k_\beta\\
       k        & -m      & -N &  0 & -N/2\\
       m        &  k      & 0  & -N &  0\\
       k_\alpha &-k_\beta &-N/2& 0  & -\frac{N(2 N -1)}{6(N-1)}
      \end{bmatrix}
\end{equation}

\begin{equation}
    \label{eq:vec_b}
     \vec{b} = 
    \begin{bmatrix}
        b_5 & b_6 & b_7 & b_8 & -k_\mu
    \end{bmatrix}^T
\end{equation}

and

\begin{equation}
\label{eq:27}
    k_\alpha:=\sum\left(\frac{i-1}{N-1}\cos \theta_i\right)
\end{equation}

\begin{equation}
\label{eq:28}
    k_\beta:=\sum\left(\frac{i-1}{N-1}\sin \theta_i\right) 
\end{equation}

\begin{equation}
\label{eq:k_mu}
    k_\mu: = \sum\left(\frac{i-1}{N-1} \xi_{P_i}\right)  
\end{equation}
The Hessian $H_S=\frac{d^2S}{d\vec{x}^2}$,
and it is non-negative definite (for the proof refer to Appendix B). 
\begin{equation*}
    \label{eq:mtx_A}
    H_S = \begin{bmatrix}
       N        &  0      & -k & -m & -k_\alpha\\
       0        &  N      & m  & -k & k_\beta\\
       -k        & m      & N &  0 & N/2\\
       -m        &  -k      & 0  & N &  0\\
       -k_\alpha &k_\beta &N/2& 0  & \frac{N(2 N -1)}{6(N-1)}
      \end{bmatrix},
\end{equation*}

Therefore, the last condition supplies the minimum of the function $S$, if $\det H_S \neq 0$.


For analytical solution in this case we refer to Eq. (\ref{eq:14.1}) -- (\ref{eq:14.4}):
\begin{equation}
    \label{eq:40}
    x_1 = \frac{b_5+k x_3+m x_4+k_\alpha x_5}{N}
\end{equation}

\begin{equation}
    \label{eq:41}
    x_2 = \frac{b_6-m x_3 k x_4 - k_\beta x_5}{N}
\end{equation}

\begin{equation}
    \label{eq:42}
    x_3 = \frac{N b_7+\frac{N^2}{2} x_5-k b_5 -k k_\alpha x_5 +m b_6 -m k_\beta x_5}{k^2 +m^2 -N^2}
\end{equation}

\begin{equation}
    \label{eq:43}
    x_4 = \frac{N b_8-m b_5 - m k_\alpha x_5 -k b_6+k k_\beta x_5}{k^2 + m^2-N^2},
\end{equation}
assuming $k^2+m^2\neq N^2$; and $x_5$ can be found directly by substituting Eq. (\ref{eq:40}) -- (\ref{eq:43}) into Eq. (\ref{eq:26}). 

\section{Additional Synthesis Conditions and Multiple Optimization Criteria}
\label{sec:additionalCrit}

In order to choose the acceptable solutions from a trial table, which contains great amount of data, a certain design criteria will be used. The trajectory of the foot center $P$, called a step cycle, consists of two phases: ``support phase" and ``transfer phase" (swing). During the support phase the foot center $P$ should trace horizontal straight-line. The \textit {main criteria} of synthesis is the accuracy of straight line generation during support phase, that has to be minimized:

\begin{equation}
    \label{eq:21w}
    c_1 = \max_{i=\overline{1,N}} |\eta_P^0-\eta_{P_i}|.
\end{equation}

At the same time we deal with a complicated synthesis task since the following additional criteria should be taken into account.

\begin{itemize}
    \item 	The chassis height $H=-x_4$ has to be minimized or (which is the same condition) $\eta^0=-H$ to be maximized:

    \begin{equation}
        \label{eq:22w}
        c_2=x_4
    \end{equation}
    
    \item	The worse transference angle has to be maximized

    \begin{equation}
        \label{eq:23w}
        c_3 = \min_{0\leq \varphi\leq 2 \Pi} (\mu_{BCD},\mu_{EFG})
    \end{equation}  

    where
    
    \begin{equation}
    \label{eq:24w}
    \begin{split}
        \mu_{BCD} &= \arccos{\frac{L_{BC}^2+L_{CD}^2-|BD|^2}{2L_{BC} L_{CD}}}, \\
        \mu_{EFG} &= \arccos{\frac{L_{EF}^2+L_{FG}^2-|EG|^2}{2 L_{EF}L_{FG}}},
    \end{split}
    \end{equation}

    $L_{BC},L_{CD}, L_{FG},L_{EF}$ are lengths of corresponding links, $|BD|, |EG|$ are distances between the centers of corresponding joints.  
    
    \item 	Anatomy matching: hip to shin relation $L_{FP}/L_{FG}$ (the ratio of the thigh to the lower leg) has to be around one ($c_4 = 1$)
    \begin{equation}
        \label{eq:26w}
        c_4=L_{FP}/L_{FG}
    \end{equation}

    \item Te penalty function $c_5$ for external transfer of the foot center $P$ has to be minimized. 
    \begin{equation}
        \label{eq:27w}
        c_5 = \sum_{i=1}^N\max((\eta^0-\eta_{P_i}),0)
    \end{equation}
\end{itemize}

If $\eta_{P_i}<\eta^0$, then the penalty function $c_5$ will be increased to positive value $\max((\eta^0-\eta_{P_i},0)$. Otherwise, if $\eta_{P_i}\leq \eta^0$, then $\max((\eta^0-\eta_{P_i}),0)=0$, thus nothing will be added to the function $c_5$. 


\section{Global Search and Multi-Criteria Optimization of Exoskeleton Link Dimensions}

\label{sec:globalsearch}

Let us keep notation $\xi, \eta$ for absolute coordinates of joints as it was in the previous sections. 

$n=13$ variable parameters of the exoskeleton mechanism  are varied within a given search area by using so called random LP-$\tau$ sequences, evenly distributed in $n$-dimensional parallelepiped \cite{statnikov1999multicriteria, ibrayev2002approximate, Ibrayev2014}.
Global search carried out specifying the following limits on variable parameters

\begin{align*}
    -1.50 &\leq \eta_D \leq 1.50 \\
    -1.50 &\leq \xi_G \leq 1.50 \\
    -1.50 &\leq \eta_G \leq 1.50 \\
    0.05 &\leq r_{AB} \leq 0.30 \\
    0^\circ &\leq \phi_0 \leq 360^\circ \\
    180^\circ &\leq \Phi \leq 190^\circ \\
    0.40 &\leq L_{BC} \leq 1.00 \\
    0.40 &\leq L_{CD} \leq 1.00 \\
    0.50 &\leq x_E \leq 1.20 \\
    -0.50 &\leq y_E \leq 1.20 \\
    0.10 &\leq L_{EF} \leq 0.40 \\
    0.30 &\leq L_{FG} \leq 1.10, \\
\end{align*}
where 

- $\xi_D, \eta_D, \xi_G, \eta_G$ are the absolute coordinates of frame joints $D$ and $G$;

- $r_{AB}$ is the length of crank $AB$;

- $x_E, y_E$ are the local coordinates of joint $E$ relative to the $Bx_2y_2$ moving coordinate system, with the $Bx_2$ axis along line $BC$;
%
%
The foot-point traces a straight line while the crank angle is in the range $\varphi_0 \leq \varphi \leq \varphi_0 + \Delta \Phi$.

The notation $p_1, p_2, ..., p_{13}$ is used in the tables for these parameters: $p_1: = \xi_D, p_2: = \eta_D, p_3:= \xi_G, p_4: = \eta_G, p_5: = r_{AB}, p_6: = \varphi_0, p_7:= \Phi, p_8: = L_{BC}, p_9:=L_{CD}, p_{10}:=x_E, p_{11}:=y_E, p_{12}:=L_{EF}, p_{13}:= L_{FG}$.

For each set of $n=13$ random variables analytical solutions for 5 design parameters $x_1, x_2,..., x_5$ are determined applying formulae Eq. (\ref{eq:40}) -- (\ref{eq:43}), and Eq. (\ref{eq:26}), and the trial table is obtained by calculating design criteria values. Analyzing the obtained trial table, 25  preliminary solutions are selected as specified in Table \ref{tab:1} and shown in Fig. \ref{fig:2} (\ref{sec:appx_C}). The criteria values ($c_1, ..., c_5$) for the obtained solutions are found to be vary within the following limits:

\begin{equation}
\label{eq:28w}
\begin{gathered}
    0 \leq c_1 \leq 0.05; \quad
    -1.6 \leq c_2 \leq -0.70; \quad
    27 \leq c_3 \leq 90; \\
    0.5 \leq c_4 \leq 2.0; \quad
    0 \leq c_5 \leq 15.00
\end{gathered}
\end{equation}

\begin{table}[htbp]
\centering
\caption{Trial table fragment: criteria values with the limits specified by  Eq. (\ref{eq:28w})}
\label{tab:1}
\begin{tabular}{ccccccc}
LP$\tau$ \textnumero & Fig & \textit{\textbf{$c_1$}} & \textit{\textbf{$c_2$}} & \textit{\textbf{$c_3$}} & \textit{\textbf{$c_4$}} & \textit{\textbf{$c_5$}} \\ 
\hline
19597            & \ref{fig:2a}          & 0,01248              & -147,377             & 3,624,502            & 0,72546              & 0,81125              \\ 
20108            & \ref{fig:2b}           & 0,01147              & -158,420             & 5,885,124            & 108,176              & 416,844              \\ 
4327             & \ref{fig:2c}           & 0,01502              & -137,035             & 3,793,790            & 0,83849              & 0,73034              \\ 
23379            & \ref{fig:2d}           & 0,01254              & -150,056             & 3,116,345            & 0,79369              & 114,581              \\ 
16038            & \ref{fig:2e}           & 0,01413              & -156,780             & 4,855,631            & 0,70010              & 225,781              \\ 
17217            & \ref{fig:2f}           & 0,01566              & -158,656             & 4,388,126            & 0,49570              & 0,98881              \\ 
15985            & \ref{fig:2j}           & 0,01930              & -146,579             & 3,042,169            & 0,98595              & 246,731              \\ 
18709            & \ref{fig:2h}           & 0,02863              & -152,807             & 3,897,082            & 0,51939              & 114,727              \\ 
25950            & \ref{fig:2i}           & 0,02381              & -129,038             & 5,426,018            & 162,689              & 818,859              \\ 
12502            & \ref{fig:2j}           & 0,02398              & -147,632             & 3,400,994            & 0,63502              & 653,650              \\ 
12709            & \ref{fig:2k}           & 0,03239              & -139,157             & 2,736,399            & 117,269              & 833,609              \\ 
4149             & \ref{fig:2l}           & 0,03486              & -130,577             & 4,095,673            & 115,044              & 904,692              \\ 
7934             &              & 0,01531              & -137,623             & 2,599,608            & 102,889              & 263,405              \\ 
25906            &              & 0,01791              & -148,873             & 5,359,325            & 197,100              & 608,409              \\ 
26153            &              & 0,01821              & -145,573             & 3,598,747            & 0,85130              & 368,757              \\ 
16074            &              & 0,02117              & -157,979             & 3,669,121            & 0,76626              & 317,949              \\ 
4838             &              & 0,02171              & - 1,32{]}063         & 3,955,225            & 0,99599              & 694,072              \\ 
29001            &              & 0,02532              & -131,698             & 2,974,295            & 168,845              & 350,371              \\ 
11761            &              & 0,02588              & -151,677             & 3,248,405            & 0,68656              & 113,792              \\ 
379              &              & 0,02937              & -111,082             & 2,736,639            & 185,211              & 455,336              \\ 
12658            &              & 0,02942              & -155,801             & 4,313,619            & 167,947              & 789,940              \\ 
15834            &              & 0,03027              & -146,200             & 5,063,574            & 160,647              & 689,696              \\ 
7260             &              & 0,03258              & -131,509             & 4,130,343            & 192,072              & 1,026,527            \\ 
32037            &              & 0,03713              & -144,800             & 3,883,024            & 124,053              & 838,340              \\ 
\end{tabular}
\end{table}

The parameter values for solution on Fig. \ref{fig:2a} are shown in Tables \ref{tab:2}-\ref{tab:4} (Appendix C)


\textbf{Global Search within the Narrowed Search Area}. 
Analyzing the results, we study functionality of the mechanism. The most elusive was meeting criteria $c_5$ (achieving internal swing with trajectory turned inward). Thus, a design criterion $c_6 $ was introduced to increase the step height $h_i = \eta_{P_i} - \eta^0$ defined as 
\begin{equation}\label{eq: c_6}
    c_6 = \sum_{i=1}^N h_i,
\end{equation}
that has to be maximized. (Note that we have not used $h = \max h_i $ as a design criterion $c_6$, since it does not reflect the foot trajectory that goes below the limit $\eta = \eta^0$ as demonstrated in Fig. \ref{fig:2} (Appendix C). Meanwhile, when using the sum these trajectories will have negative sign which decreases the sum).
Arranging and cutting the obtained trial table, eliminating solutions with unacceptable criteria values, we are looking for the compromise solutions that meet all designing criteria, so we clarify new boundaries for design variables. Then we carry out new search of the design parameters within the new search area and obtain new trial table. After several repetition of this sequence of actions we came to the following search area specified by the boundaries:

\begin{gather*}
    -0.77 \leq \xi_D \leq -0.05 \\
    -0.84 \leq \eta_D \leq 0.05 \\
    -0.95 \leq \xi_G \leq -0.06 \\
    -0.91 \leq \eta_G \leq 0.04 \\
    0.07 \leq r_{AB} \leq 0.22 \\
    -110^\circ \leq \varphi_0 \leq -9^\circ \\
    180^\circ \leq \Phi \leq 190^\circ \\
    0.20 \leq L_{BC} \leq 0.75 \\
    0.19 \leq L_{CD} \leq 0.60 \\
    0.80 \leq x_E \leq 2.22 \\
    -0.70 \leq y_E \leq 1.07 \\
    0.17 \leq L_{EF} \leq 0.62 \\
    0.57 \leq L_{FG} \leq 1.91. \\
\end{gather*}

\begin{figure}[h!]
    \centering
    \begin{subfigure}{0.24\textwidth}
        \centering
        \includegraphics[height=5cm]{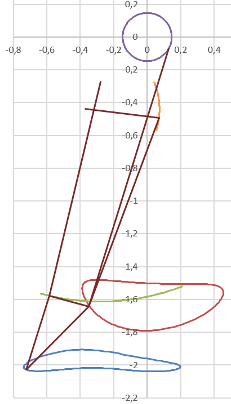}
        \caption{}
        \label{fig:21a}
    \end{subfigure}
    \begin{subfigure}{0.24\textwidth}
        \centering
        \includegraphics[height=5cm]{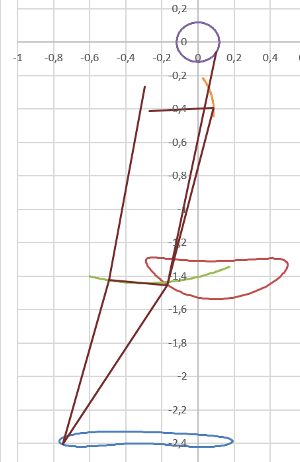}
        \caption{}
        \label{fig:21b}
    \end{subfigure}
    \begin{subfigure}{0.24\textwidth}
        \centering
        \includegraphics[height=5cm]{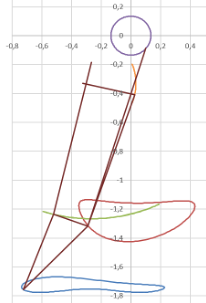}
        \caption{}
        \label{fig:21c}
    \end{subfigure}
    \begin{subfigure}{0.24\textwidth}
        \centering
        \includegraphics[height=5cm]{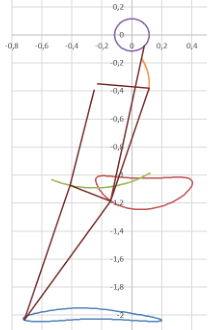}
        \caption{}
        \label{fig:21d}
    \end{subfigure}
    \caption{Visualization of the mechanisms with swing phase trajectories turned inward}
    \label{fig:21}
\end{figure}

As the result we obtained a number of solutions presented on Table \ref{tab:21}, having internal transfer segment of the foot, so that in the swing phase the foot trajectory is turned inward. Part of these solutions are plotted on Fig. \ref{fig:21}. The solutions in Table \ref{tab:21} are arranged  by criterion $c_6$ in descending order. Despite the step height (height of the foot transference) not being very high, we obtained the desired solutions with high accuracy (about 1 percent from step stride) and fine transmission angle (from $32^\circ$ to $55^\circ$) (Tables \ref{tab:22} -- \ref{tab:24}).

\begin{table}[htbp]
\centering
\caption{The trial table fragment: the best solutions by criterion $c_6$}
\label{tab:21}
\begin{tabular}{cccccccc}
\small
LP$\tau$ \textnumero & Fig. & \textit{\textbf{$c_1$}} & \textit{\textbf{$c_2$}} & \textit{\textbf{$c_3$}} & \textit{\textbf{$c_4$}} & \textit{\textbf{$c_5$}} & \textit{\textbf{$c_6$}} \\
\hline
29884            & \ref{fig:21a}          & \textbf{0,011}          & -2,028                  & \textbf{32,8}          & 0,351                  & 0,332                  & 5,504                  \\
31076            & \ref{fig:21b}          & \textbf{0,011}          & -2,411                  & \textbf{54,8}          & \textbf{0,865}         & 0,329                  & 4,617                  \\
26230            & \ref{fig:21c}          & 0,012                  & \textbf{-1,764}         & \textbf{40,1}          & 0,516                  & 0,423                  & 4,435                  \\
27664            &              & \textbf{0,011}          & -2,549                  & \textbf{48,1}          & 0,435                  & 0,331                  & 4,293                  \\
1832             &              & \textbf{0,008}          & -2,605                  & \textbf{37,1}          & \textbf{1,008}         & 0,299                  & 3,773                  \\
7592             &              & \textbf{0,010}          & -2,595                  & \textbf{49,7}          & 0,520                  & 0,277                  & 3,617                  \\
17728            &              & \textbf{0,010}          & -2,627                  & \textbf{54,4}          & 0,714                  & 0,340                  & 3,571                  \\
7146             & \ref{fig:21d}          & \textbf{0,009}          & -2,036                  & \textbf{35,4}          & \textbf{1,441}         & 0,342                  & 3,546                  \\
15805            &              & 0,012                  & \textbf{-1,805}         & \textbf{33,5}          & 0,639                  & 0,431                  & 3,218                  \\
15409            &              & 0,013                  & \textbf{-1,830}         & \textbf{34,1}          & \textbf{1,169}         & 0,468                  & 3,169                  \\
14440            &              & \textbf{0,009}          & -2,133                  & 27,0                   & \textbf{1,503}         & 0,327                  & 2,990                  \\
17650            &              & \textbf{0,010}          & -2,246                  & \textbf{53,0}          & 0,642                  & 0,310                  & 2,360                  \\
9674             &              & \textbf{0,008}          & -1,957                  & \textbf{52,7}          & \textbf{1,017}         & 0,265                  & 1,925                  \\
30047            &              & \textbf{0,011}          & \textbf{-1,547}         & \textbf{44,2}          & 0,589                  & 0,508                  & 1,512                  \\
11346            &              & 0,012                  & \textbf{-1,800}         & \textbf{38,7}          & \textbf{0,811}         & 0,426                  & 1,423                  \\
30486            &              & \textbf{0,011}          & -1,959                  & \textbf{35,9}          & 0,670                  & 0,563                  & 1,256                  \\
8344             &              & \textbf{0,010}          & -1,922                  & \textbf{32,6}          & 2,057                  & 1,130                  & 1,137                  \\
20277            &              & 0,012                  & \textbf{-1,712}         & \textbf{39,7}          & 0,533                  & 0,842                  & 1,126                  \\
380              &              & 0,013                  & \textbf{-1,794}         & \textbf{48,2}          & 0,641                  & 0,395                  & 1,045                  \\
10474            &              & 0,013                  & \textbf{-1,808}         & \textbf{32,5}          & 0,789                  & 0,711                  & 1,019                  \\
16159            &              & 0,013                  & \textbf{-1,314}         & 27,4                   & 0,513                  & 2,275                  & 0,892                  \\
15009            &              & 0,013                  & -1,909                  & \textbf{34,4}          & 0,754                  & 0,819                  & 0,858                  \\
2630             &              & \textbf{0,010}          & -1,922                  & \textbf{34,6}          & 0,793                  & 0,393                  & 0,779                  \\
27197            &              & 0,012                  & \textbf{-1,468}         & \textbf{32,1}          & \textbf{0,840}         & 0,673                  & 0,775                  \\
13726            &              & 0,012                  & \textbf{-1,641}         & \textbf{38,8}          & \textbf{1,614}         & 0,696                  & 0,730                  \\
18746            &              & \textbf{0,011}          & -1,914                  & \textbf{46,9}          & 0,794                  & 0,384                  & 0,522                 
\end{tabular}
\end{table}

\begin{table}[h!]
\centering
\caption{Solution \textit{7146} design parameters $p_1, ..., p_7$}
\label{tab:22}
\small
\begin{tabular}{cccccccc}
\begin{tabular}[c]{@{}c@{}}LP$\tau$ \\ \textnumero\end{tabular} & \textit{\textbf{$X_D$}} & \textit{\textbf{$Y_D$}} & \textit{\textbf{$X_G$}} & \textit{\textbf{$Y_G$}} & \textit{\textbf{$r_{AB}$}} & \textit{\textbf{$\varphi_0$}} & \textit{\textbf{$\Phi$}} \\
\hline
7146             & -0,22953             & -0,34842             & -0,25044             & -0,39308             & 0,11482               & -43,16928            & 185,16479          
\end{tabular}
\end{table}

\begin{table}[h!]
\centering
\caption{Solution \textit{7146} design parameters $p_8, \ldots, p_{13}$}
\label{tab:23}
\begin{tabular}{ccccccc}
LP$\tau$ \textnumero & \textit{L\textsubscript{BC}} & \textit{L\textsubscript{CD}} & \textit{x\textsubscript{E}} & \textit{y\textsubscript{E}} & \textit{L\textsubscript{EF}} & \textit{L\textsubscript{FG}} \\
\hline
7146 & 0.30241 & 0.34472 & 1.08084 & -0.33612 & 0.29188 & 0.69900
\end{tabular}
\end{table}

\begin{table}[h!]
\centering
\caption{Solution \textit{7146} design parameters parameters $x_1, x_2, ..., x_5$}
\label{tab:24}
\centering
\begin{tabular}{cccccc}
LP$\tau$ \textnumero & \textit{\textbf{$x_1$}} & \textit{\textbf{$x_2$}} & \textit{\textbf{$x_3$}} & \textit{\textbf{$x_4$}} & \textit{\textbf{$x_5$}} \\
\hline
7146             & 0.19607              & 1.00248              & -0.71623             & -2.03609             & 0.90610             
\end{tabular}
\end{table}



One can observe that solutions \textit{29884} (Fig. \ref{fig:21a}) and \textit{26230} (Fig. \ref{fig:21c}) exhibit a low value of criterion $c_4$, leading to the displacement of the knee joint $E$ to an undesirable lower position that does not conform to human anatomy. Solution \textit{31076} (Fig. \ref{fig:21b}) possesses an acceptable $c_4$ value, but the straight-line segment height $H=-c_2 = -\eta^0$ is excessive. Same shortcomings take place in solutions \textit{27664}, \textit{1832}, \textit{7592}, and \textit{17728}, thus the relevant figures have not been plotted. Ultimately, our analysis identified solution \textit{7146} (Fig. \ref{fig:21d}) as the most suitable choice. Although it exhibits a suboptimal swing height, it has excellent accuracy, transmission angle, and a satisfactory knee joint position $E$. As demonstrated in Fig. \ref{fig:21}, the lower the $c_6$ value, the lower the foot swing height ($h$). Thus, the rest of the solutions are not shown.

\section{Local Search and Final Mechanism Design}
\label{sec:finaldesign}
In order to improve the swing height the local search around the solution \textit{7146} is carried out. The selected solutions presented on Table \ref{tab:25},  arranged in descending order of the criterion $c_6$. 

\begin{figure}[H]
    \centering
    \begin{subfigure}{0.25\textwidth}
        \centering
        \includegraphics[width=\linewidth]{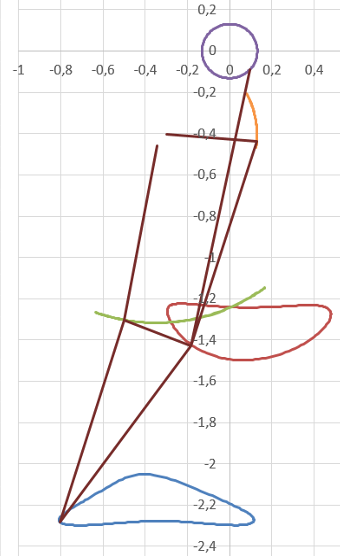}
        \caption{}
        \label{fig:22a}
    \end{subfigure}
    \begin{subfigure}{0.25\textwidth}
        \centering
        \includegraphics[width=\linewidth]{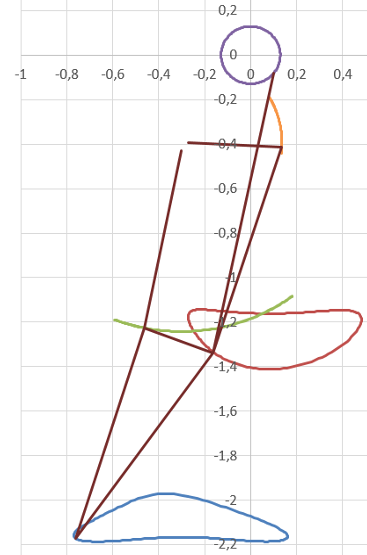}
        \caption{}
        \label{fig:22b}
    \end{subfigure}
    \begin{subfigure}{0.25\textwidth}
        \centering
        \includegraphics[width=\linewidth]{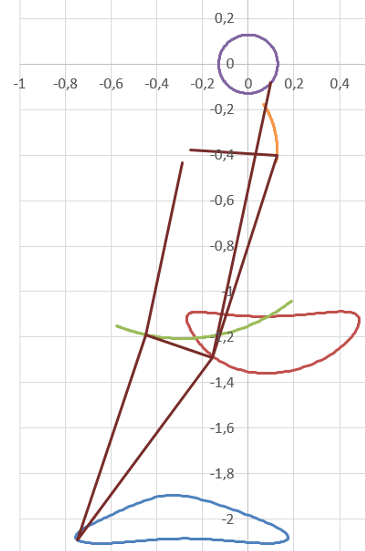}
        \caption{}
        \label{fig:22c}
    \end{subfigure}
    \begin{subfigure}{0.25\textwidth}
        \centering
        \includegraphics[width=\linewidth]{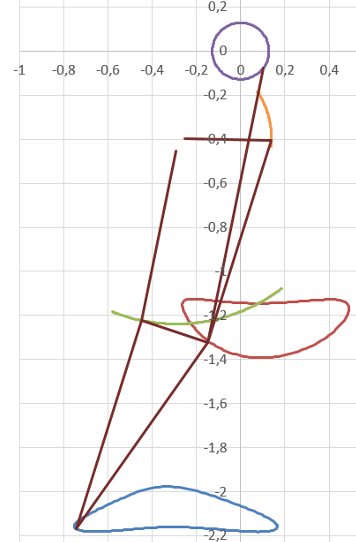}
        \caption{}
        \label{fig:22d}
    \end{subfigure}
    \begin{subfigure}{0.25\textwidth}
        \centering
        \includegraphics[width=\linewidth]{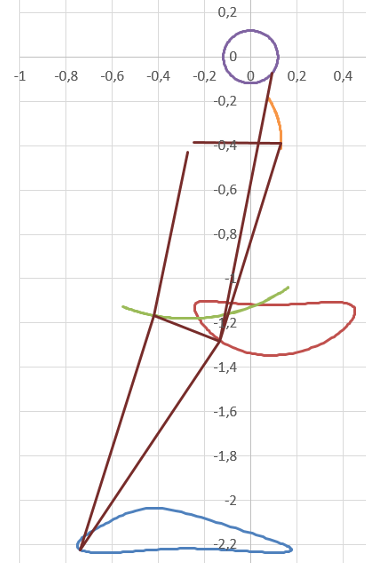}
        \caption{}
        \label{fig:22e}
    \end{subfigure}
    \begin{subfigure}{0.25\textwidth}
        \centering
        \includegraphics[width=\linewidth]{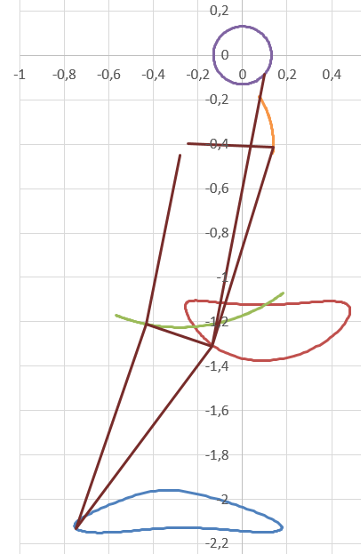}
        \caption{}
        \label{fig:22f}
    \end{subfigure}
    \caption{The results by the local search with improved swing height}
    \label{fig:22}
\end{figure}

Because of the substantial number of solutions, we included only few images on Fig. \ref{fig:22}, Table \ref{tab:26} explains why we have selected these specific images. The remaining solutions are not depicted, since one can observe, that step height decreases. 
The design parameters for the solution \textit{8398} are presented on Tables \ref{tab:27} -- \ref{tab:29}.

\begin{table}[H]
\centering
\caption{The results of the local search around the solution \textit{7146}: the best solutions by criterion $c_6$}
\label{tab:25}
\begin{tabular}{cccccccc}
LP$\tau$ \textnumero & Fig. & $c_1$ & $c_2$ & $c_3$ & $c_4$ & $c_5$ & $c_6$ \\
\hline
5380 & \ref{fig:22a} & $\mathbf{0.010}$ & $-2.288$ & $20.6$ & $1.199$ & $0.347$ & $\mathbf{9.385}$ \\
20757 & & $\mathbf{0.011}$ & $\mathbf{-2.101}$ & $22.2$ & $1.278$ & $0.389$ & $\mathbf{8.880}$ \\
16365 & & $\mathbf{0.011}$ & $-2.112$ & $23.6$ & $1.100$ & $0.373$ & $\mathbf{8.817}$ \\
22170 & & $\mathbf{0.011}$ & $-2.156$ & $22.1$ & $1.427$ & $0.351$ & $\mathbf{8.361}$ \\
17370 & \ref{fig:22b} & $\mathbf{0.010}$ & $-2.176$ & $23.7$ & $1.220$ & $0.346$ & $\mathbf{8.337}$ \\
29005 & \ref{fig:22c} & $\mathbf{0.011}$ & $\mathbf{-2.096}$ & $25.3$ & $1.228$ & $0.372$ & $\mathbf{8.242}$ \\
19111 & & $\mathbf{0.011}$ & $\mathbf{-2.031}$ & $22.0$ & $1.317$ & $0.390$ & $\mathbf{8.161}$ \\
26668 & & $\mathbf{0.010}$ & $-2.223$ & $24.3$ & $1.210$ & $0.367$ & $\mathbf{8.070}$ \\
13257 & & $\mathbf{0.011}$ & $-2.110$ & $23.3$ & $1.415$ & $0.331$ & $\mathbf{8.060}$ \\
17228 & & $\mathbf{0.011}$ & $-2.201$ & $22.1$ & $1.502$ & $0.334$ & $\mathbf{8.053}$ \\
17743 & & $\mathbf{0.012}$ & $\mathbf{-1.972}$ & $20.2$ & $1.409$ & $0.368$ & $\mathbf{8.052}$ \\
19215 & & $\mathbf{0.012}$ & $\mathbf{-1.992}$ & $22.1$ & $1.327$ & $0.383$ & $\mathbf{7.986}$ \\
8398 & \ref{fig:22d} & $\mathbf{0.011}$ & $-2.171$ & $\mathbf{26.9}$ & $1.260$ & $0.376$ & $\mathbf{7.934}$ \\
3766 & & $\mathbf{0.010}$ & $-2.173$ & $23.7$ & $1.342$ & $0.343$ & $\mathbf{7.924}$ \\
23411 & & $\mathbf{0.012}$ & $\mathbf{-2.066}$ & $\mathbf{27.1}$ & $1.331$ & $0.377$ & $\mathbf{7.920}$ \\
30022 & & $\mathbf{0.011}$ & $-2.158$ & $24.6$ & $1.278$ & $0.418$ & $\mathbf{7.913}$ \\
12749 & & $\mathbf{0.011}$ & $\mathbf{-2.096}$ & $\mathbf{26.2}$ & $1.358$ & $0.347$ & $\mathbf{7.861}$ \\
30915 & & $\mathbf{0.011}$ & $\mathbf{-2.033}$ & $22.2$ & $1.276$ & $0.384$ & $\mathbf{7.769}$ \\
1031 & & $\mathbf{0.011}$ & $\mathbf{-2.071}$ & $\mathbf{28.4}$ & $1.230$ & $0.389$ & $\mathbf{7.749}$ \\
3628 & & $\mathbf{0.011}$ & $-2.215$ & $25.9$ & $1.235$ & $0.402$ & $\mathbf{7.727}$ \\
3327 & & $\mathbf{0.012}$ & $\mathbf{-1.985}$ & $22.3$ & $1.224$ & $0.428$ & $\mathbf{7.704}$ \\
10981 & & $\mathbf{0.011}$ & $\mathbf{-2.078}$ & $23.0$ & $1.310$ & $0.395$ & $\mathbf{7.680}$ \\
10134 & & $\mathbf{0.010}$ & $-2.165$ & $24.6$ & $1.171$ & $0.364$ & $\mathbf{7.665}$ \\
4693 & & $\mathbf{0.011}$ & $\mathbf{-2.077}$ & $22.4$ & $1.165$ & $0.422$ & $\mathbf{7.663}$ \\
2872 & \ref{fig:22e} & $\mathbf{0.0098}$ & $-2.227$ & $24.1$ & $1.466$ & $0.312$ & $\mathbf{7.630}$ \\
20824 & & $\mathbf{0.011}$ & $-2.282$ & $23.4$ & $1.044$ & $0.364$ & $\mathbf{7.619}$ \\
31 & & $\mathbf{0.010}$ & $\mathbf{-2.030}$ & $22.7$ & $1.090$ & $0.374$ & $\mathbf{7.618}$ \\
15153 & & $\mathbf{0.010}$ & $-2.111$ & $23.0$ & $1.335$ & $0.384$ & $\mathbf{7.617}$ \\
28345 & \ref{fig:22f} & $\mathbf{0.011}$ & $-2.138$ & $\mathbf{29.2}$ & $1.255$ & $0.363$ & $\mathbf{7.603}$ \\
\end{tabular}
\end{table}

\begin{table}[H]
\centering
\caption{Comments to the images}
\label{tab:26}
\begin{tabular}{p{1.5cm}p{1.5cm}p{9cm}}
LP$\tau$ \textnumero & Fig. & Comments                                                                                                              \\
\hline
5380    & \ref{fig:22a}  & The best solution by swing height, the accuracy is high                                                               \\
17370   & \ref{fig:22b}  & The best solution by the accuracy, step height is high                                                                \\
29005   & \ref{fig:22c}  & Transmission angle and height H are better than in previous solutions, swing height and the accuracy are high as well \\
8398    & \ref{fig:22d}  & Trying to improve transmission angle we come to this solution                                                         \\
2872    & \ref{fig:22e}  & The best accuracy 0.0098                                                                                              \\
28345   & \ref{fig:22f}  & The best transmission angle 29.2                                                                                     
\end{tabular}
\end{table}

\begin{table}[H]
\centering
\caption{Design parameters $p_1, \ldots, p_7$ for solution \textit{8398}}
\label{tab:27}
\small
\begin{tabular}{@{}cccccccc@{}}
\begin{tabular}[c]{@{}c@{}}LP$\tau$ \\ \textnumero\end{tabular} & $\mathit{X_D}$ & $\mathit{Y_D}$ & $\mathit{X_G}$ & $\mathit{Y_G}$ & $\mathit{r_{AB}}$ & $\mathit{\varphi_o}$ & $\mathit{\Phi}$ \\
\hline
8398 & -0.25122 & -0.39679 & -0.29140 & -0.45175 & 0.12829 & -37.87341 & 183.5455 \\
\end{tabular}
\end{table}

\begin{table}[H]
\centering
\caption{Design parameters $p_8, \ldots, p_{13}$ for solution \textit{8398}}
\label{tab:28}
\begin{tabular}{ccccccc}
LP$\tau$ \textnumero & $L_{BC}$ & $L_{CD}$ & $x_E$ & $y_E$ & $L_{EF}$ & $L_{FG}$ \\
\hline
8398 & 0.32738 & 0.39084 & 1.20469 & -0.39171 & 0.31884 & 0.78648 \\

\end{tabular}
\end{table}

\begin{table}[H]
\centering
\caption{Design parameters $x_1, x_2, \ldots, x_5$ for solution 8398}
\label{tab:29}
\begin{tabular}{ccccccc}
LP$\tau$ \textnumero  & ${x_1}$ & ${x_2}$ & ${x_3}$ & ${x_4}$ & ${x_5}$ \\
\hline
8398 & 0.30476 & 0.99082 & -0.74396 & -2.17071 & 0.90412 \\
\end{tabular}
\end{table}

\section{Conclusions}
\label{sec:conclusions}

The typical schemes of leg exoskeletons are based on open-loop kinematic chain with the motors mounted directly on the moveable joints. While the design choice offer greater flexibility and ease of design, their large number of  DOF contributes to increased costs and complexities in control. Using heavy servo-motors to meet significant torques aroused in active joints leads to complicated and cumbersome design. Existing literature emphasizes bulkiness and substantial weight of this kind of devices.

Another approach involves the utilization of 1-DOF mechanisms. However, many of them characterized by a substantial number of linkages, often reaching eight or more. For example, the best schemes of such design are selected in figure below. As you can see in Fig. \ref{fig:conclusion}, even the mechanism (b), which is often considered the best and published in the MMT International Journal as one of the respectable designs \cite{desai2019analysis}, is overly bulky with a large number of links also. If employed as an lower-limb exoskeleton mechanism, its geometry would not align with the human anatomy. In the figure also shown that the mechanism we designed (a) is well-suited to the anatomical parameters of humans. Note, that the numbers on the axes are relative values (not in meters), when L --- the step length --- is equal to 1.

In this study, we introduced a novel synthesis method with analytical solutions provided for synthesizing lower-limb exoskeleton. Additionally, we have incorporated multicriteria optimization by six designing criteria. As a result, we offer several mechanisms, comprising only six links, well-suited to the human anatomical structure, exhibit superior trajectory accuracy, efficient force transmission, satisfactory step height, and having internal transfer segment of the foot. 

\begin{figure}[htbp]
    \centering
    
    \begin{subfigure}{0.125\textwidth}
        \centering
        \includegraphics[width=\linewidth]{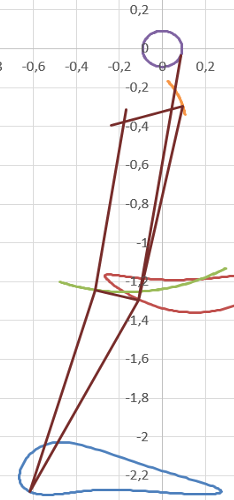}
        \caption{}
        \label{fig:conclusion_a}
    \end{subfigure}
    \begin{subfigure}{0.2\textwidth}
        \centering
        \includegraphics[width=\linewidth]{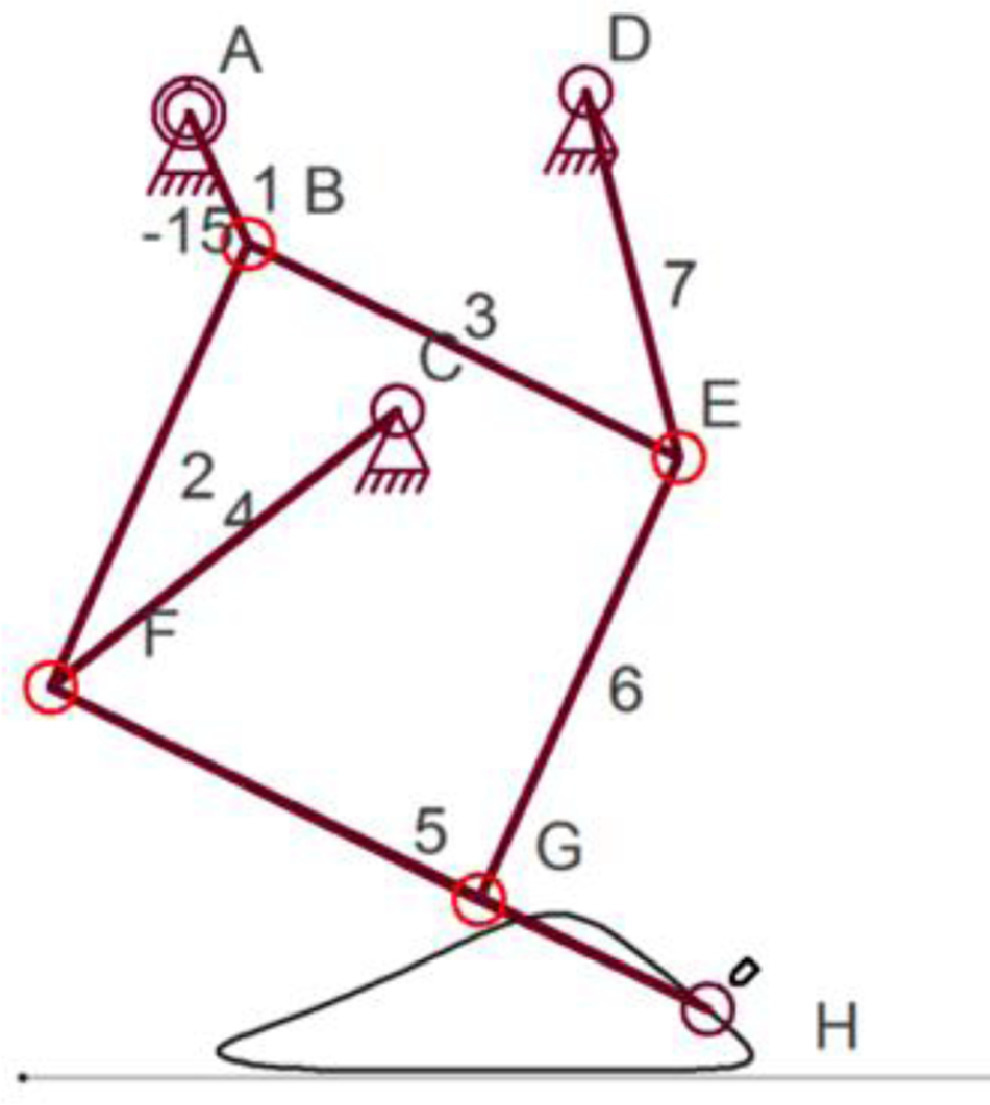}
        \caption{}
        \label{fig:conclusion_b}
    \end{subfigure}
    \begin{subfigure}{0.2\textwidth}
        \centering
        \includegraphics[width=\linewidth]{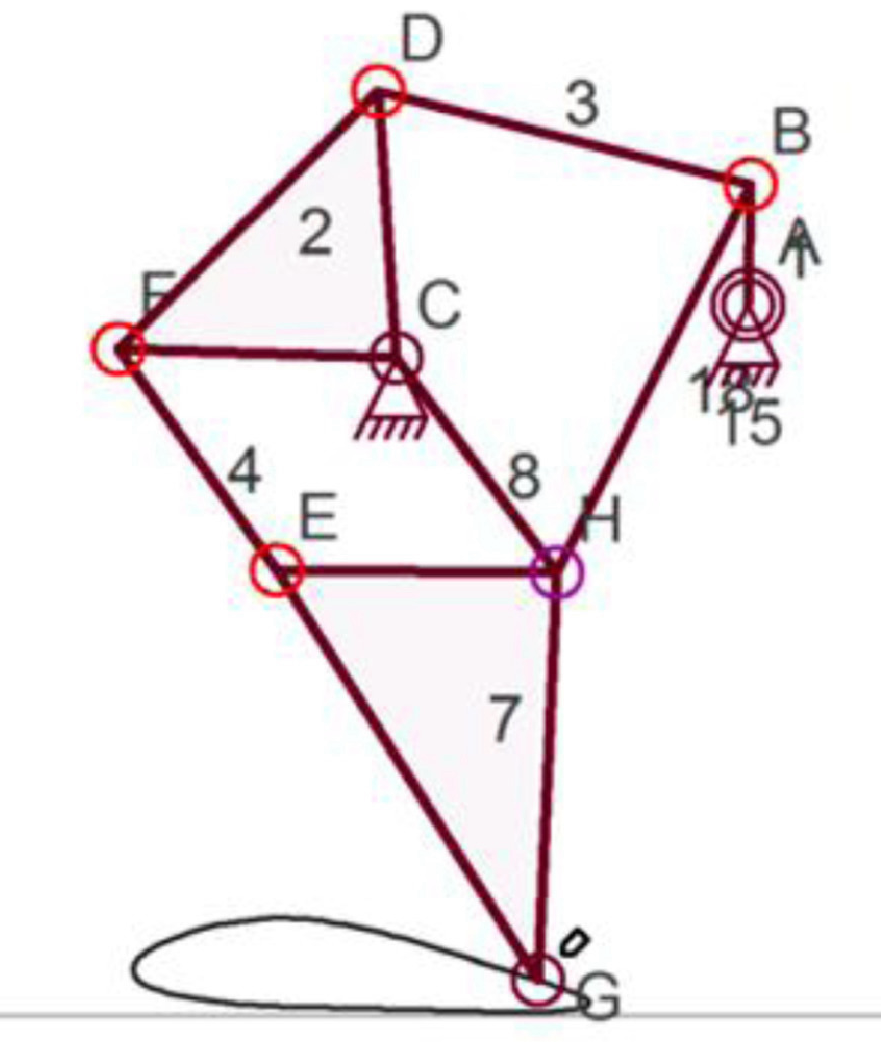}
        \caption{}
        \label{fig:conclusion_c}
    \end{subfigure}
    \begin{subfigure}{0.2\textwidth}
        \centering
        \includegraphics[width=\linewidth]{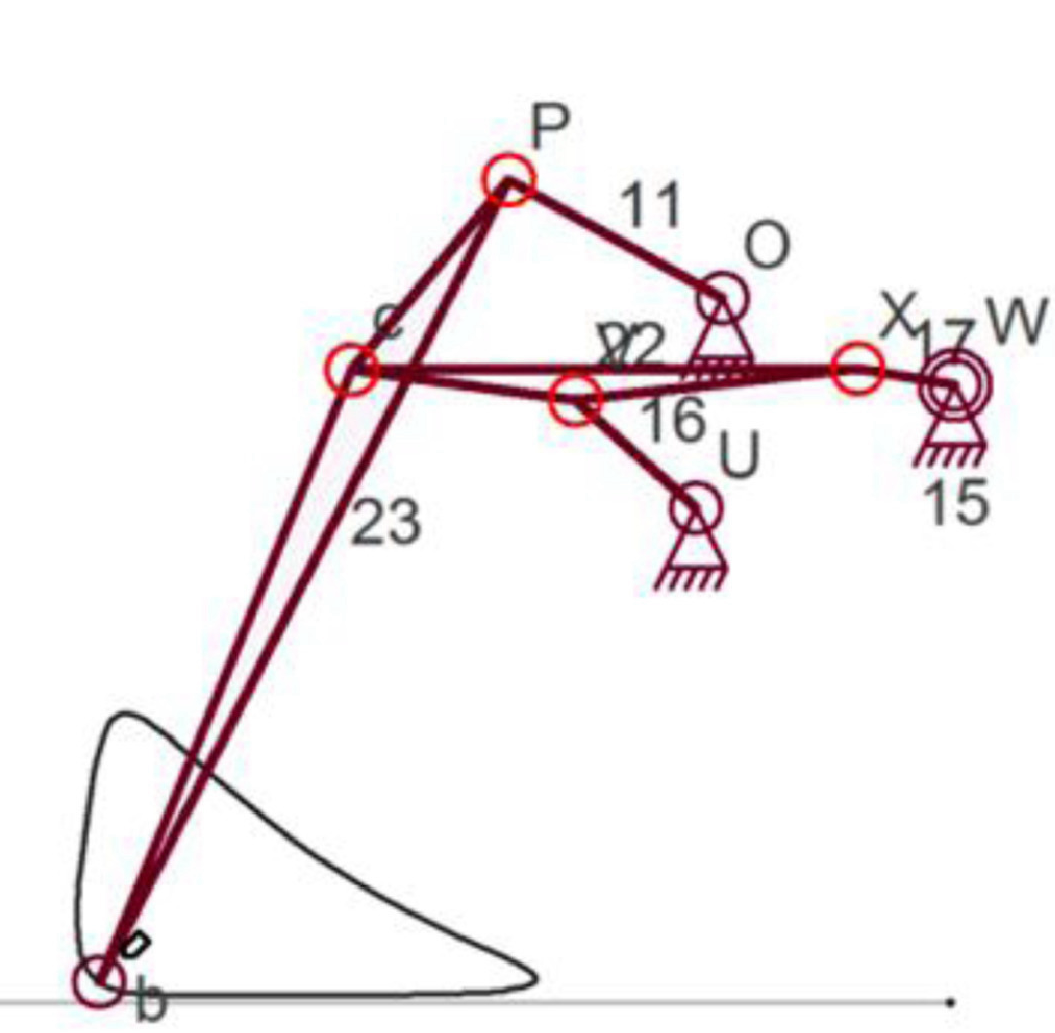}
        \caption{}
        \label{fig:conclusion_d}
    \end{subfigure}
    \centering
    \caption{a) Our 6-bar mechanism; b) 8-bar Peaucellier-Lipkin type mechanism \cite{desai2019analysis}; c) Theo Jansen's Linkage \cite{desai2019analysis}; d) Joseph Klann's Linkage \cite{desai2019analysis}}
    \label{fig:conclusion}
   
\end{figure}

\begin{acknowledgment}
This work was supported by the Science Committee of the Ministry of Education and Science of the Republic of Kazakhstan under Grant AP14870080 "Structural-Parametric Synthesis of the Musculoskeletal  Mechanisms of the Exoskeleton of the lower limb". The authors would like to express their gratitude King Abdullah University of Science and Technology for their partial financial support.
\end{acknowledgment}

%
% ...



\appendix       
\section*{Appendix A: Kinematics of the Lower Limb Exoskeleton Mechanism}
\label{sec:appendixA}
\begin{figure}[h!]
    \centering
    \begin{subfigure}{0.3\textwidth}
        \centering
        \includegraphics[width=\linewidth]{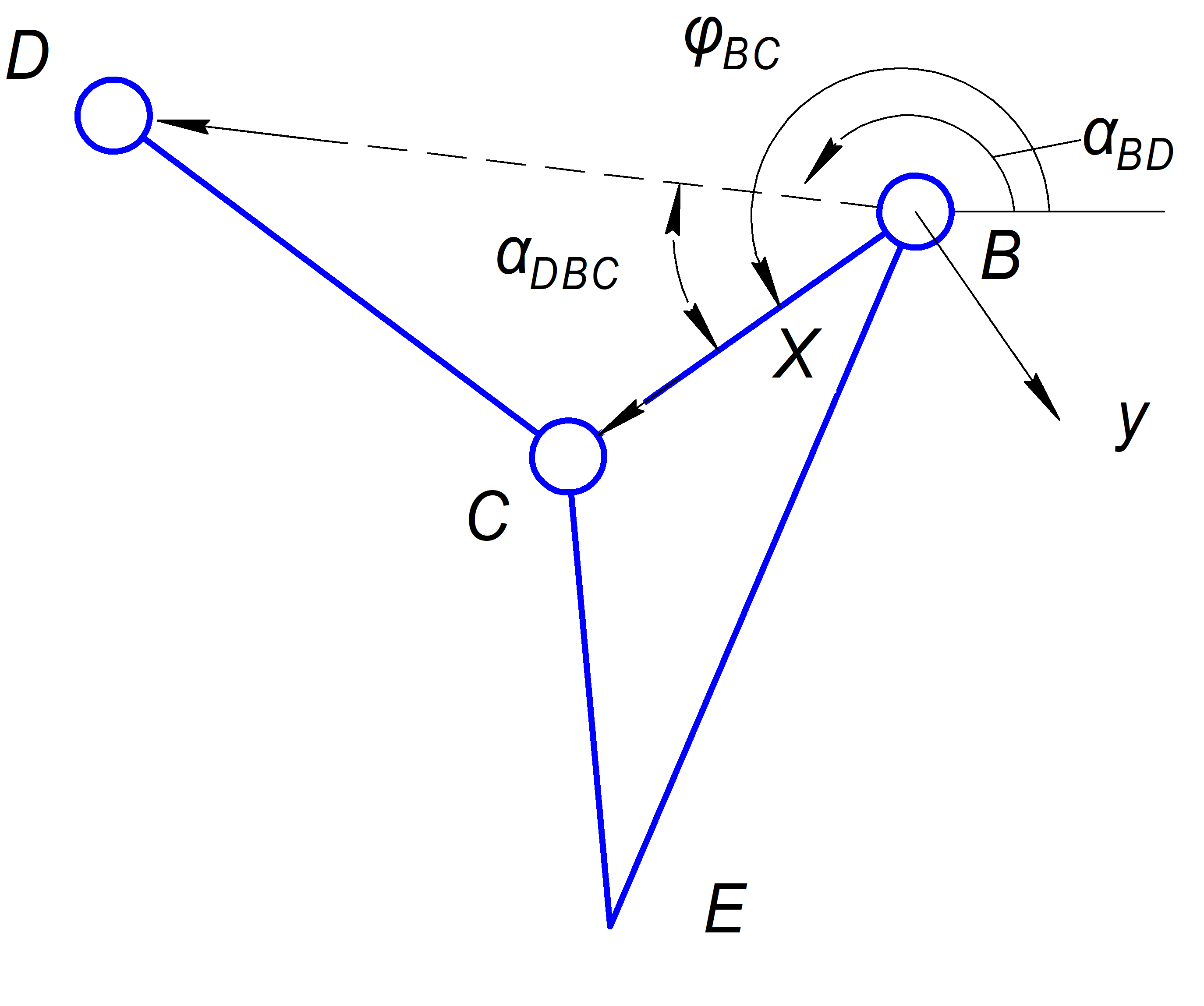}
        \caption{}
        \label{fig:BDCE}
    \end{subfigure}
    \hspace{1.5cm}
    \begin{subfigure}{0.22\textwidth}
        \centering
        \includegraphics[width=\linewidth]{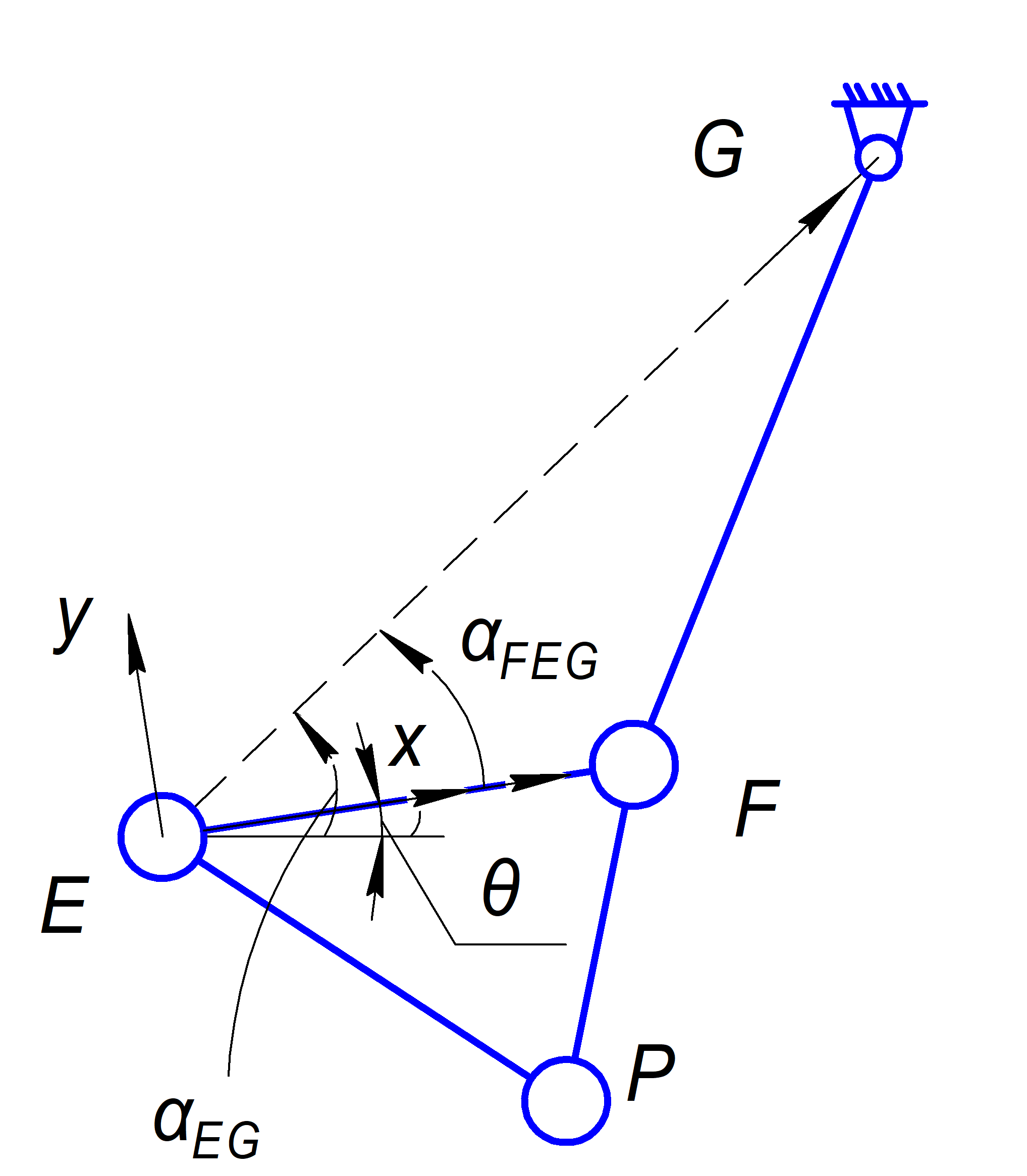}
        \caption{}
        \label{fig:GEFB}
    \end{subfigure}
    \hfill
    \caption{Kinematic scheme of a lower-limb exoskeleton mechanism}
    \label{fig:angles}
\end{figure}

The absolute coordinates of joints $B$ and $C$ are simply determined as follows: 
\begin{equation}
    \label{eq:2:2}
    \begin{aligned}[c]
        \xi_B &= \xi_A + r_{AB} \cos{\varphi}\\
        \eta_B &= \eta_A + r_{AB} \sin{\varphi}
    \end{aligned}
\end{equation}
\begin{equation}
    \label{eq:2:3}
    \begin{aligned}[c]
        \xi_C &= \xi_B + l_{BC} \cos{\varphi_{BC}}\\
        \eta_C &= \eta_B + l_{BC} \sin{\varphi_{BC}},
    \end{aligned}
\end{equation}
where the angular position $\varphi_{BC} = \angle( \scalebox{0.8}{$\vec{A\xi}$}, \scalebox{0.8}{$\vec{BC}$} )$ (Fig. \ref{fig:angles}) of link $BC$ is defined as 
 
\begin{equation}
    \varphi_{BC} = \alpha_{BD} + \alpha_{DBC},
\end{equation}

\begin{equation}
    \alpha_{BD} = \angle ( \scalebox{0.8}{$\overrightarrow{A\xi}$}, \scalebox{0.8}{$\overrightarrow{BD}$} ) = \operatorname{atan2} (\xi_D-\xi_B, \eta_D-\eta_B),
\end{equation}

\begin{equation}
    \alpha_{DBC} = \arccos{\frac{|BD|^2+l_{BC}^2-l_{CD}^2}{2 l_{BC} |BD|}},
\end{equation}

\begin{equation}
    |BD|^2 = (\xi_D-\xi_B)^2+ (\eta_D-\eta_B)^2.
\end{equation}

The absolute coordinates of the joint $E$ are

\begin{equation}
    \label{eq:E}
    \begin{aligned}[c]
        \xi_E &= \xi_B + x_E \cos{\varphi_{BC}}-y_E \sin{\varphi_{BC}}\\
        \eta_E &= \eta_B + x_E \sin{\varphi_{BC}}+y_E \cos{\varphi_{BC}},
    \end{aligned}
\end{equation}
where $x_E, y_E$ are local coordinates of joint $E$ relative to $Bxy$. Now, in the same way we analyse the dyad $EFG$, and determine the absolute coordinates of joint $F$:

\begin{equation}
    \label{eq:F}
    \begin{aligned}[c]
        \xi_F &= \xi_E + l_{EF} \cos{\varphi_{EF}}\\
        \eta_F &= \eta_E + l_{EF} \sin{\varphi_{EF}},
    \end{aligned}
\end{equation}
where $\varphi_{EF} = \angle( \scalebox{0.8}{$\vec{A\xi}$}, \scalebox{0.8}{$\vec{EF}$} )$ is determined as:

\begin{equation}
    \varphi_{EF} = \alpha_{EG} + \alpha_{FEG}
\end{equation}

\begin{equation}
    \alpha_{EG} = \angle ( \scalebox{0.8}{$\overrightarrow{A\xi}$}, \scalebox{0.8}{$\overrightarrow{EG}$} ) = \operatorname{atan2} (\xi_G-\xi_E, \eta_G-\eta_E)
\end{equation}

\begin{equation}
    \alpha_{FEG} = \arccos{\frac{|EG|^2+l_{EF}^2-l_{FG}^2}{2 l_{EF} |EG|}}
\end{equation}

\begin{equation}
    |EG|^2 = (\xi_G-\xi_E)^2+ (\eta_G-\eta_E)^2.
\end{equation}

The equations for the foot-center $P$ are given in the Section \ref{sec:eqs}.

\section*{Appendix B: Proof of the Optimization Condition}
\label{sec:appx_B}



The original $2N$ constraint equations ($\delta_i^\xi =0, \delta_i^\eta=0$) can be written in the form:
\begin{equation}
\begin{aligned}
\delta_i^\xi=0: \quad & \cos \theta_i x_1 - \sin \theta_i x_2 -x_3 - \frac{i-1}{N-1}x_5 +\xi_{E_i}=0, \quad i=\overline{1,N} \\
\delta_i^\eta=0: \quad & \sin \theta_i x_1 + \cos\theta_i x_2 - x_4 - \eta_{E_i} =0.
\end{aligned}
\end{equation}

Let us write these equations in the matrix form:
\begin{equation}
    \begin{bmatrix}
        \Gamma_i & | -T_i
    \end{bmatrix} \vec{x}
    =
    \begin{bmatrix}
        -\xi_{E_i} \\
        -\eta_{E_i}
    \end{bmatrix},
\end{equation}
where

\begin{equation}
    \Gamma_i = 
    \begin{bmatrix}
        \cos \theta_i & -\sin \theta_i \\
        \sin \theta_i & \cos \theta_i
    \end{bmatrix},
\end{equation}

\begin{equation}
  T_i = 
    \begin{bmatrix}
        1 & 0 & \frac{i-1}{N-1}\\
        0 & 1 & 0
    \end{bmatrix},
\end{equation}
\begin{equation}
    \dim \Gamma_i = 2 \times 2, \quad \dim T_i = 2 \times 3.
\end{equation}

Then the linear system of equations ($5 \times 5$) for determining $\vec x$ can be written in the form
\begin{equation}
    H_1^T H_1 \vec x=H_1^T \vec c_1,
\end{equation}
where $H_1$ is the matrix of dimension $\dim H_1 = 2N \times 5, \dim \vec c_1 = 2N$:
\begin{equation}
    H_1= \begin{bmatrix}
        \Gamma_1 & -T_1\\
        \Gamma_2 & -T_2\\
        \vdots & \vdots\\
        \Gamma_N & -T_N
    \end{bmatrix}
\end{equation}
\begin{equation}
    c_1 = \begin{bmatrix}
        \xi_{E_1}\\
        \eta_{E_1}\\
        \xi_{E_2}\\
        \eta_{E_2}\\
        \vdots\\
        \xi_{E_N}\\
        \eta_{E_N}.
    \end{bmatrix}
\end{equation}

Thus, the Hessian matrix $H_S$ will be 
\begin{equation}
    \frac{d^2 S}{d\vec x ^2}= H_1^T H_1
\end{equation}

Therefore, the matrix $H_S$ is non-negative definite, $\det H_S = 0 $ is the singularity of the synthesis problem. 

\section*{Appendix C}
 \label{sec:appx_C}

\begin{figure}[t]
    \centering
    \begin{subfigure}{0.23\textwidth}
        \centering
        \includegraphics[height=4cm]{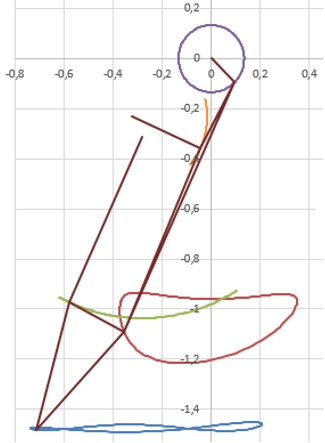}
        \caption{}
        \label{fig:2a}
    \end{subfigure}
    \hspace{0.05em}
    \begin{subfigure}{0.23\textwidth}
        \centering
        \includegraphics[height=4cm]{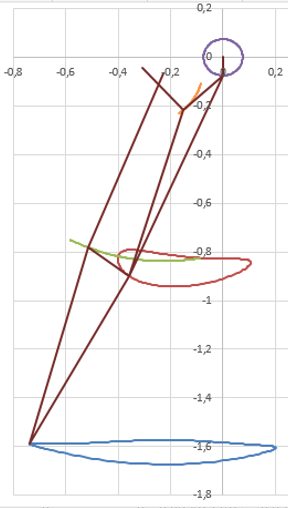}
        \caption{}
        \label{fig:2b}
    \end{subfigure}
    \hspace{0.05em}
    \begin{subfigure}{0.23\textwidth}
        \centering
        \includegraphics[height=4cm]{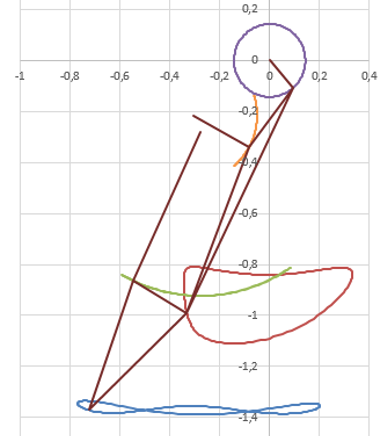}
        \caption{}
        \label{fig:2c}
    \end{subfigure}
    \hspace{0.05em}
    \begin{subfigure}{0.23\textwidth}
        \centering
        \includegraphics[height=4cm]{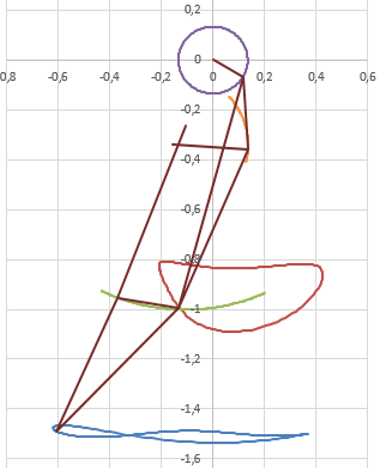}
        \caption{}
        \label{fig:2d}
    \end{subfigure}


    \begin{subfigure}{0.23\textwidth}
        \centering
        \includegraphics[height=4cm]{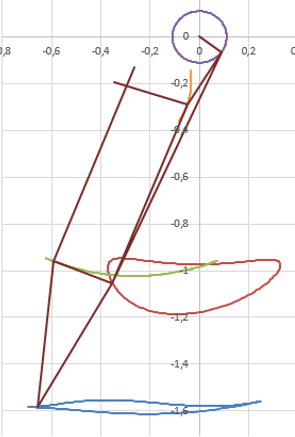}
        \caption{}
        \label{fig:2e}
    \end{subfigure}
    \hspace{0.05em}
    \begin{subfigure}{0.23\textwidth}
        \centering
        \includegraphics[height=4cm]{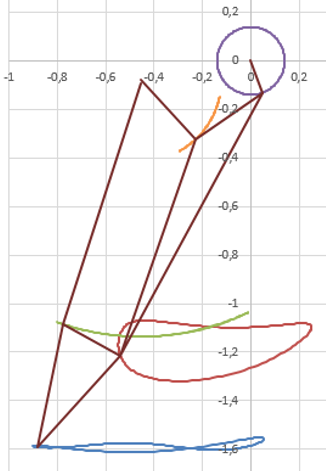}
        \caption{}
        \label{fig:2f}
    \end{subfigure}
    \hspace{0.05em}
    \begin{subfigure}{0.23\textwidth}
        \centering
        \includegraphics[height=4cm]{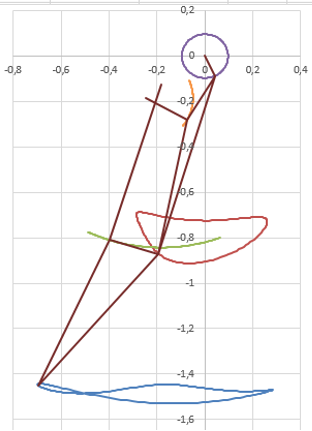}
        \caption{}
        \label{fig:2g}
    \end{subfigure}
    \hspace{0.05em}
    \begin{subfigure}{0.23\textwidth}
        \centering
        \includegraphics[height=4cm]{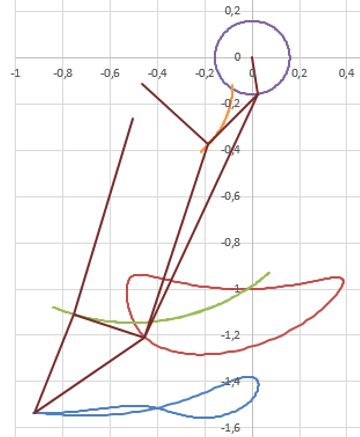}
        \caption{}
        \label{fig:2h}
    \end{subfigure}


    \begin{subfigure}{0.23\textwidth}
        \centering
        \includegraphics[height=4cm]{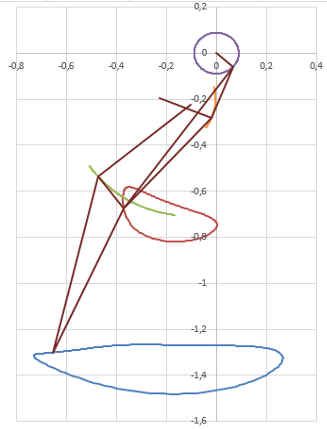}
        \caption{}
        \label{fig:2i}
    \end{subfigure}
    \hspace{0.05em}
    \begin{subfigure}{0.23\textwidth}
        \centering
        \includegraphics[height=4cm]{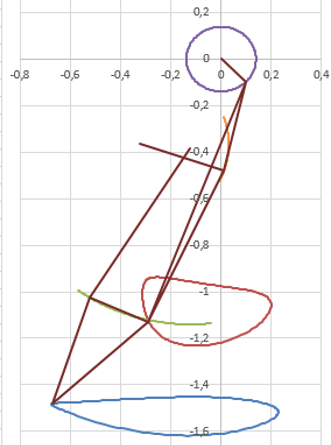}
        \caption{}
        \label{fig:2j}
    \end{subfigure}
    \hspace{0.05em}
    \begin{subfigure}{0.23\textwidth}
        \centering
        \includegraphics[height=4cm]{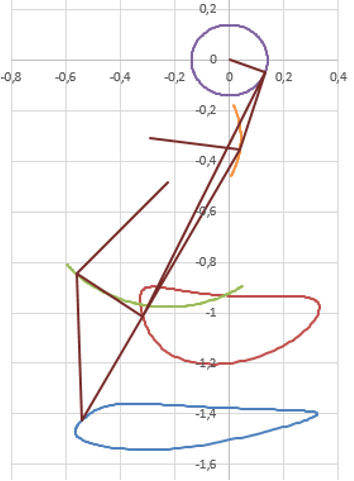}
        \caption{}
        \label{fig:2k}
    \end{subfigure}
    \hspace{0.05em}
    \begin{subfigure}{0.23\textwidth}
        \centering
        \includegraphics[height=4cm]{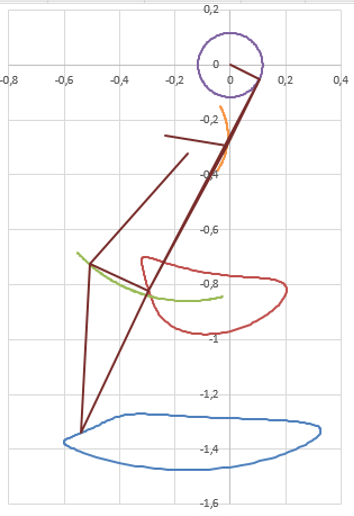}
        \caption{}
        \label{fig:2l}
    \end{subfigure}

    \caption{Functionality study of 6 bar mechanism}
    \label{fig:2}
\end{figure}


\begin{table}[H]
\centering
\caption{Random search design parameters $p_1,...,p_7$ for the mechanism on Fig. \ref{fig:2a}}
\label{tab:2}
\small
\setlength{\tabcolsep}{4pt} 
\begin{tabular}{cccccccc}

LP$\tau$  \textnumero  & \textit{\textbf{$\xi_D$}} & \textit{\textbf{$\eta_D$}} & \textit{\textbf{$\xi_G$}} & \textit{\textbf{$\eta_G$}} & \textit{\textbf{$r_{AB}$}} & \textit{\textbf{$\varphi_0$}} & \textit{\textbf{$\Phi$}} \\
\hline
19597            & -0,32133             & -0,23007             & -0,28094             & -0,31117             & 0,13375               & -45,44359            & 183,12408 \\

\end{tabular}
\end{table}


\begin{table}[H]
\centering
\caption{Random search design parameter values $p_8, \ldots, p_{13}$ for the mechanism on the Fig. \ref{fig:2a}}
\label{tab:3}
\begin{tabular}{ccccccc}
LP$\tau$ \textnumero & \textbf{$L_{BC}$} & \textbf{$L_{CD}$} & \textbf{$x_E$} & \textbf{$y_E$} & \textbf{$L_{EF}$} & \textbf{$L_{FG}$} \\
\hline
19597            & 0.29670      & 0.30422               & 1.09131              & 0.06663              & 0.25049               & 0.72561       
\end{tabular}
\end{table}

\begin{table}[H]
\centering
\caption{The parameters $x_1, x_2, \ldots, x_5$ for the mechanism on the Fig. \ref{fig:2a}}
\label{tab:4}
\begin{tabular}{ccccccc}
LP$\tau$ \textnumero & Fig & \textbf{$x_1$} & \textbf{$x_2$} & \textbf{$x_3$} & \textbf{$x_4$} & \textbf{$x_5$} \\
\hline
19597            & \ref{fig:2a}           & 0.12676              & 0.51165              & -0.71130             & -1.47377             & 0.91107              \\            
\end{tabular}
\end{table}

\end{document}